\pdfoutput=1

\documentclass[11pt]{article}
\usepackage[table]{xcolor}
\usepackage[]{acl}

\usepackage{times}
\usepackage{latexsym}

\usepackage[T1]{fontenc}

\usepackage[utf8]{inputenc}

\usepackage{microtype}
\usepackage{inconsolata}
\usepackage{multirow}
\usepackage{arydshln}
\usepackage{subcaption}

\usepackage{booktabs}
\usepackage{amsmath}
\usepackage{cuted,tcolorbox,lipsum}

\NewTColorBox{NewBox}{ s O{} O{!htbp}  }{%
  floatplacement={#3},
  IfBooleanTF={#1}{float*,width=\textwidth}{float},
  colframe=blue!50!black,colback=blue!10!white,
  title={#2},
  }

\newcommand{\selfrag}[1]{\texttt{SELF-RAG}}
\newcommand{\selfragmult}[1]{\texttt{SELF}-\textit{multi}-\texttt{RAG}}
\newcommand{\selfragorg}[1]{\texttt{SELF-RAG$_{s}$}}

\newlength{\Oldarrayrulewidth}

\title{Learning When to Retrieve, What to Rewrite, \\and How to Respond in Conversational QA}

\author{
\begin{minipage}[t]{\textwidth}
\centering
Nirmal Roy$^{\dag}$\thanks{\hspace{0.2cm}Work done as an intern at Amazon.} \hspace{0.1cm}, Leonardo F. R. Ribeiro$^{\ddag}$, Rexhina Blloshmi$^{\ddag}$,
Kevin Small$^{\ddag}$\vspace{1mm} 
\end{minipage}
\\
\rule{0pt}{2.5ex}
  $^{\dag}$TU Delft, $^{\ddag}$Amazon \\
  {\small \texttt{nroy000@gmail.com}, \texttt{\{leonribe, blloshmi, smakevin\}@amazon.com}} \\
}

\begin{document}
{\makeatletter\acl@finalcopytrue
  \maketitle
}
\begin{abstract}

Augmenting Large Language Models (LLMs) with information retrieval capabilities (i.e., Retrieval-Augmented Generation (RAG)) has proven beneficial for knowledge-intensive tasks. However, understanding users' contextual search intent when generating responses is an understudied topic for conversational question answering (QA). This conversational extension leads to additional concerns when compared to single-turn QA as it is more challenging for systems to comprehend conversational context and manage retrieved passages over multiple turns. In this work, we propose a method for enabling LLMs to decide \textit{when} to retrieve in RAG settings given a conversational context. When retrieval is deemed necessary, the LLM then rewrites the conversation for passage retrieval and judges the relevance of returned passages before response generation. Operationally, we build on the single-turn \selfrag{} framework~\cite{asai2023self} and propose \selfragmult{} for conversational settings. \selfragmult{} demonstrates improved capabilities over single-turn variants with respect to retrieving relevant passages (by using summarized conversational context) and assessing the quality of generated responses. Experiments on three conversational QA datasets validate the enhanced response generation capabilities of \selfragmult{}, with improvements of $\sim$13\% 
measured by human annotation.

\end{abstract}

\nopagebreak
\section{Introduction}
\label{sec:introduction}
\begin{figure}[!h]
    \includegraphics[width=\linewidth]{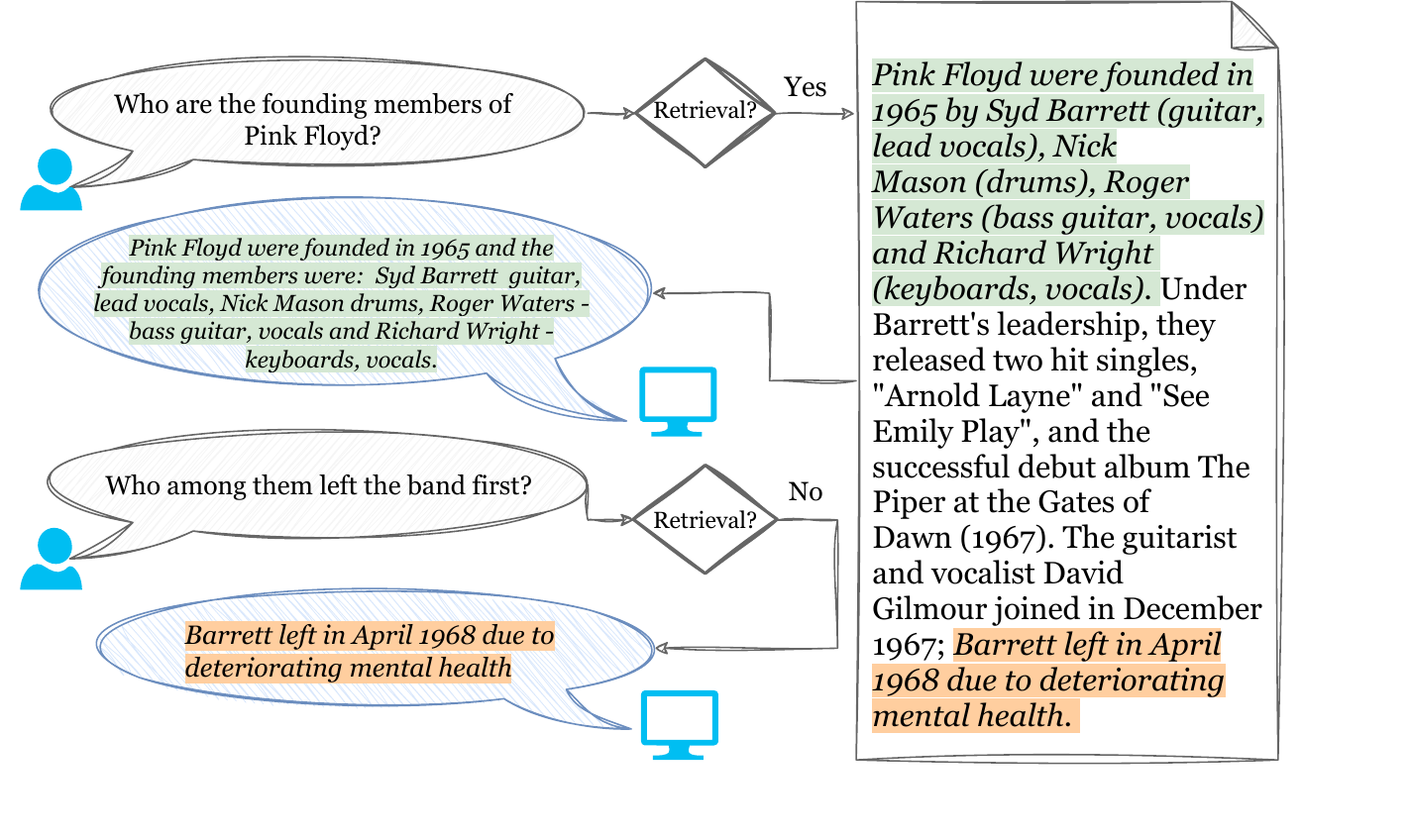} 
    \caption{\textbf{Understanding conversational context.} In multi-turn conversations user questions often refer to responses in previous turns based on passages already retrieved, as shown in the example above. To answer the follow-up question, it is not necessary to retrieve new passages and the LLM should refer back to the previously retrieved passage, which contains the response.} 
    \label{fig:singlevsmulti}
  \end{figure}
Recent advances in LLM technology has made the conversational search paradigm~\cite{culpepper2018research} increasingly viable for mainstream use as next generation search technology.
Unlike traditional search engines which primarily process keyword queries, users can treat conversational search systems as a knowledgeable expert and directly engage in a multi-turn natural language conversation to better resolve their search needs. However, despite impressive abilities in a variety of tasks including response generation and conversational understanding, factual errors and hallucinations remain persistent problems for LLM-based systems~\cite{mallen2023not,wei2024longform}. 

Retrieval-Augmented Generation (RAG) methods have been shown to partially ameliorate these issues by augmenting the input of LLMs with relevant retrieved passages, aiming to reduce factual errors in knowledge-intensive tasks~\cite{lewis2020retrieval}. Nonetheless, these approaches also can impede the flexibility of LLMs, introducing extraneous or unrelated passages or providing conflicting information with previous context/turns~\cite{adlakha-etal-2022-topiocqa}, resulting in low-quality generation~\cite{shi2023large}. Specifically, retrieving passages indiscriminately, without considering whether the factual grounding is beneficial can compromise the quality of the generated content~\cite{shi2023large,oh2023detrimental}. 

Thus, understanding \textit{if retrieval is necessary} for high-quality response generation is an important research question, especially in the context of conversational QA. In multi-turn question answering, comprehending users' contextual intent and generating responses pose significant challenges due to complexities introduced by the extended context window containing previous user interactions~\cite{aliannejadi2020harnessing, mao2023learning, wu2024faithful}.
The system, when deciding whether to retrieve or estimating the usefulness of its own response, must process a longer context of the conversation history, understand the user intent of current turn, ensure prevention of information repetition, maintain user engagement, etc. An example is provided in Figure~\ref{fig:singlevsmulti}, which shows that the decision to retrieve or not might depend on the conversational context rather than last turn only.

\begin{figure}
    \includegraphics[width=\linewidth]{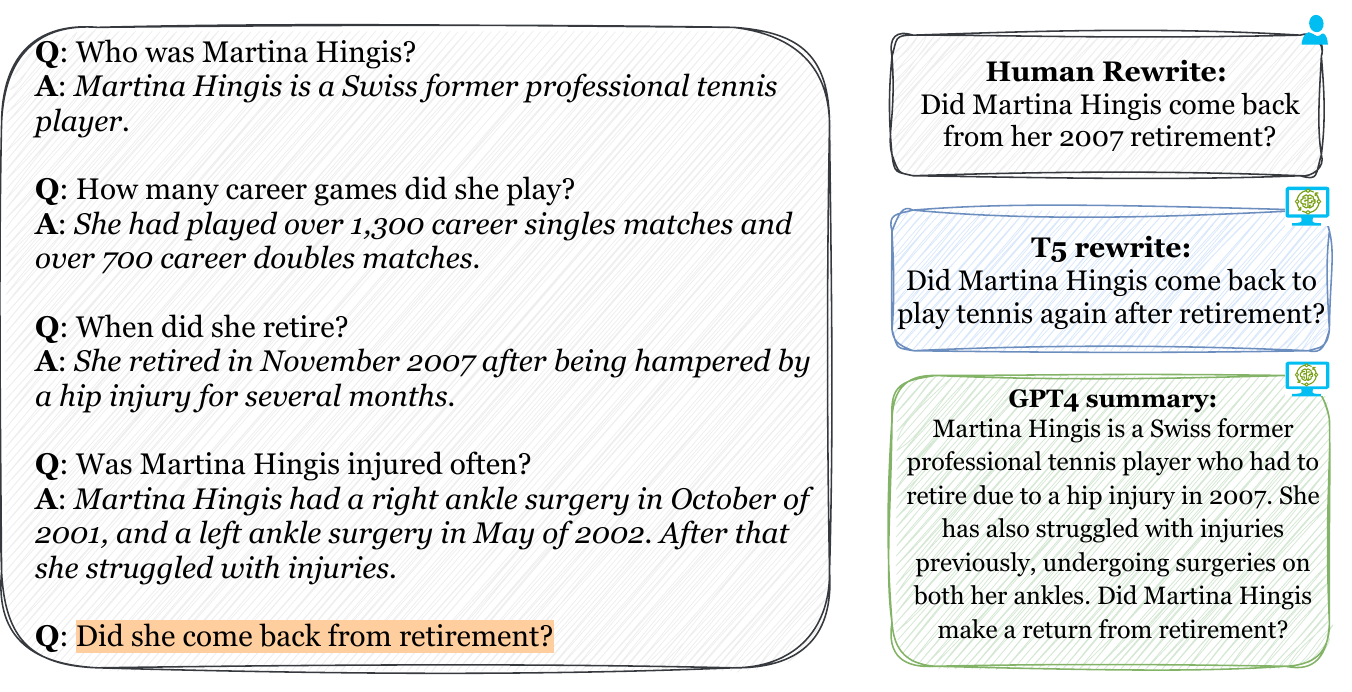} 
    \caption{\textbf{Summarizing conversational context.} While using the entire conversational context as query to a retrieval model might introduce noise, using traditional rewriting methods might miss on important aspects of the conversations. Conversation summaries provide an adaptable approach as a retrieval query.}
    \label{fig:summarymoti}
  \end{figure}

Furthermore, when retrieval is expensive or noisy, it is beneficial to utilize already retrieved documents in the conversation, given they are relevant and contain the necessary answers. Additionally, detrimental context as a result of noisy retrieval can degrade response generation quality~\cite{shi2023large,oh2023detrimental}. Lastly, when conversation history is longer, traditional conversational query rewriting methods~\cite{anantha2020open,ishii2022integrating,ye2023enhancing} that typically emphasize co-reference resolution, might be insufficient to contain all the information required for an effective retrieval~\cite{bai2024mtbench101}. As shown in Figure~\ref{fig:summarymoti}, both the gold and a T5-based query rewriting miss potentially important signals (e.g., hip injury) for retrieving correct passages. In that case, representing the conversation history in a summarized form can lead to more effective retrieval. 

In this work, we propose \selfragmult{}, an approach for efficient retrieval during a multi-turn conversation for improved response generation. \selfragmult{} provides refined contextual signals for retrieval and enhances multi-turn answer generation by potentially reusing already retrieved passages in previous turns. In particular, \selfragmult{} enables: (i) better \textit{understanding of the conversational context} to decide whether retrieval is needed and to generate useful responses accordingly in a conversational setting and (ii) \textit{summarizing the conversational context} such that it can be used as a query to retrieve relevant documents, when needed, with high retrieval effectiveness. 
Our specific contribution are as follows:
\begin{itemize}

    \item We propose \selfragmult{}, a framework to train LLMs to respond to question-answer turns by adaptively retrieving passages and reflecting on the retrieved passages in multi-turn setting. \selfragmult{} determines the necessity of retrieval given a conversational context and summarizes the conversation into a query for use with off-the-shelf  retrieval models.
    
    \item We conduct extensive experiments to observe that response quality of \selfragmult{} significantly outperforms \selfrag{} with an average improvement of $\sim$13\% for conversational datasets, measured by human annotations in Table~\ref{tab:main_results}. Moreover, \selfragmult{} summarization capabilities improves the retrieval effectiveness by 13.5\% on average (R@5), compared to query rewriting baselines (Table~\ref{tab:ret_eff}).
\end{itemize}


\section{Related Work}
\label{sec:related}

\paragraph{RAG.}
Retrieval-Augmented Generation (RAG) augments the LM input with retrieved text passages~\cite{lewis2020retrieval,guu2020retrieval}, leading to large improvements in knowledge-intensive tasks 
~\cite{ram2023context}. 
However, the improved task performance of such approaches have been shown to come at the expense of runtime efficiency~\cite{mallen2023not}, robustness to irrelevant context~\cite{oh2023detrimental,shi2023large}, and lack of attributions~\cite{liu2023evaluating,gao2023enabling}.~\citet{yoran2023making} use a natural language inference model and~\citet{xu2023recomp} employ a summarization model to filter out or compress retrieved passages before using them to prompt the LM to generate the output. In comparison, \selfrag{}~\cite{asai2023self} processes passages in parallel and filters out irrelevant ones through self-reflection, without relying on external models at inference. The self-reflection mechanism of \selfrag{} also evaluates other aspects of the model output quality, including factuality and attribution. However, \selfrag{} is not trained to comprehend conversational context, which we specifically equip \selfragmult{} to do.~\citet{kulkarni2024reinforcement} propose a reinforcement learning (RL) based approach where the policy model can perform two actions: fetch conversation context or skip retrieval. Their approach was shown to save costs by reducing tokens when the model decides retrieval is not needed, while also slightly improving response generation. In contrast, the goal of \selfragmult{} is not only to decrease retrieval redundancy but also increase retrieval effectiveness. 

\paragraph{LLMs and Multi-turn Conversations.}
In order to enable LLMs to interact with humans in a dialogue-based settings, the standard approach is to collect multi-turn instructions~\cite{chiang2023vicuna,ji2023towards}, often synthetically generated using strong LLMs, and used to fine tune the LLMs for the task of response generation. 
This process is known as instruction fine-tuning, which enables LLMs to generate responses in a multi-turn dialogue setting. LLMs have also been used to perform conversational history modeling by rewriting user question~\cite{mao2023large,ye2023enhancing,wang2023domain}. Such query rewriting using LLMs have been shown to improve effectiveness for the retrieval of grounding passages. However, none of these works explicitly train the LLMs to reflect whether retrieval is needed or not (given the conversation history) or how to deal with irrelevant passages while generating RAG responses.

\paragraph{Conversational Query Rewriting.}
Query rewriting plays a vital role in enhancing conversational search by transforming context-dependent user queries into self-contained forms. Existing approaches~\cite{wu2021conqrr,mo2023convgqr} primarily leverage human-rewritten queries as labels to train query rewriting models and typically aim to convert the conversational into a single question.~\cite{ye2023enhancing} proposed to rewrite queries using the conversation history to make more informative queries. They show that rewriting queries by prompting ChatGPT with information from context helps in more effective retrieval performance. ~\citet{kaiser2024robust} performed RL based reformulations for better retrieval of entities for conversational QA over knowledge graphs. 
Their reformulations are entity focused where the answers to the questions are entities (as compared to sentences in the case of open domain QA).
~\citet{ishii2022integrating} proposed query rewriting based on a reward based system. The current question + conversation history is passed to the QA model (e.g., RoBERTa) to extract answer span from provided evidence document.~\citet{jang2024itercqr} rely on information retrieval signals directly to perform conversational query rewriting instead of relying on human-rewritten query as supervision signal. While the above works show improved retrieval effectiveness as compared to a human rewrite of the conversational context they \textit{do not} evaluate how the retrieved passages affect response generation of the models. Furthermore, we hypothesize that summarizing the conversational context instead of rewriting them to a single question will help in improving retrieval effectiveness and consequent response generation performance.

\section{\selfragmult{}}
\label{sec:method}

In this work, we propose \selfragmult{}, which extends \selfrag{}~\cite{asai2023self} to generate responses in a conversational setting. Our proposed methodology trains a LLM to comprehend longer conversational contexts to learn when retrieval is needed and also critiquing the quality of the passages and its own generation given the previous turns by generating \textit{special tokens}. Importantly, when the LLM decides retrieval is needed to generate a response, we train it to summarize the conversational context for more effective passage retrieval, which consequently leads to better response generation. \selfragmult{} has two important functions:
\paragraph{Understanding Conversational Context.} 
In a multi-turn context, evaluating the relevance of a retrieved passage involves considering both the current question and the previous conversation history, unlike a single-turn question. Furthermore, retrieved passages from previous turns can also be provided in the conversational context or a user may simply provide a passage and ask questions based on that. When the cost of retrieval is high, a model should \textit{not} decide to retrieve if the question in the current turn can be answered from passages retrieved in previous turns (as shown in Figure~\ref{fig:singlevsmulti}). In that case, the model should not only comprehend the conversational context but also the previously retrieved passages to be able to determine the necessity of retrieval. Utilizing already retrieved passages (given they are relevant) not only mitigates the harmful effects of noisy retrieval~\cite{oh2023detrimental}, but also saves costs by reducing the number of context tokens~\cite{kulkarni2024reinforcement}.

\paragraph{Summarising Conversational Context.} Representing conversational context using a single question is difficult~\cite{anantha2020open} and might result in loss of information while retrieving relevant passages (as shown in Figure~\ref{fig:summarymoti}). This is especially true in case of long conversations. Traditional conversation query rewriting methods~\cite{ye2023enhancing,mo2023convgqr,lin2020conversational} are trained to select important parts of the conversation but, typically, in a single question format. Hence, we hypothesize that summarising a conversational context can potentially include more relevant signals when retrieving passages without adding noise. This can be beneficial for both sparse and bi-encoder based dense retrieval.

\begin{figure}
    \includegraphics[width=\linewidth]{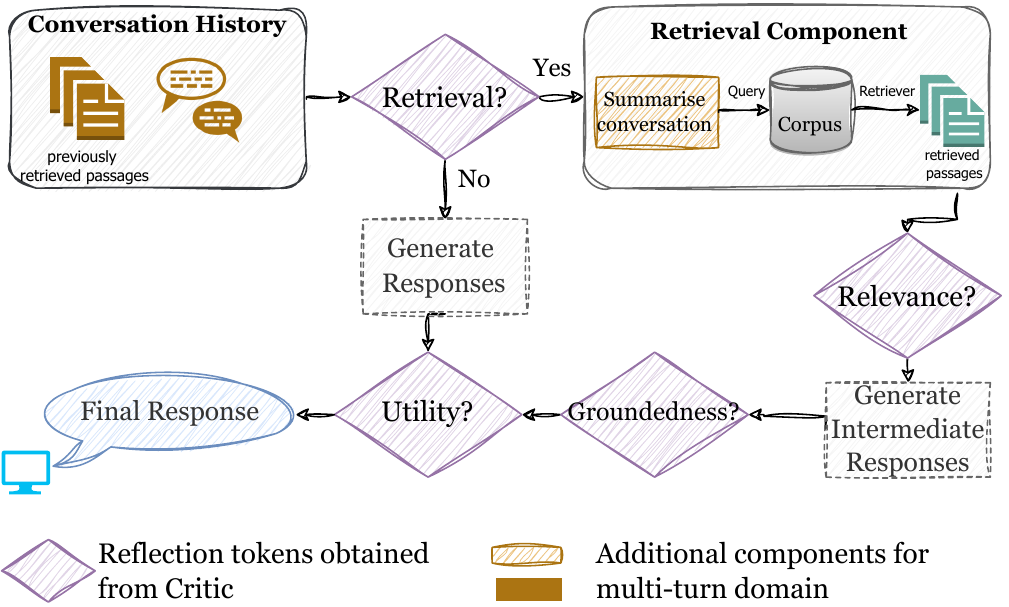} 
    \caption{\textbf{\selfragmult{} framework.} Components of the pipeline highlighted in yellow are specific to multi-turn conversations. The critic model is used to obtain the special reflection tokens that the generator model is trained to predict while generating response.}
    \label{fig:selfmultrag}
  \end{figure}


\subsection{Components}
Analogous to \selfrag{}, \selfragmult{} has three main components: (i) Critic, (ii) Generator, and (iii) Retriever. 

\paragraph{Critic.}
The task of the critic model is to output the special reflection tokens given a conversational context (as compared to a single question like in the original \selfrag{} framework). We employ the five critic tasks introduced by~\citet{asai2023self}. However, we redesign the framework to include a conversational history instead of a single-turn question.  The important distinction of our approach is that it teaches the critic model to judge whether retrieval is needed or not, and relevance of retrieved documents based on the entire conversation history. The critic tasks and special tokens are shown in Table~\ref{tab:critic_data}.

\begin{table}
\small
    \centering
    \resizebox{\columnwidth}{!}{
    \begin{tabular}{p{1.2cm}p{2.6cm}p{2.5cm}p{3.7cm}p{1cm}}
    \toprule
        \textbf{Task} & \textbf{Definition} & \textbf{\# instances}& \textbf{Tokens}& \textbf{Total}   \\ \midrule
      
      \multirow{2}{1.5cm}{\texttt{Retrieval}}  & whether retrieval is   &  \multirow{2}{2.5cm}{232; Q=96, U=136} &  [Retrieve] &  69\%  \\
         &  needed&  & [No Retrieve] &  31\%  \\ \hline
         \multirow{3}{1.5cm}{\texttt{3-way Retrieval}} &  whether retrieval is  &   \multirow{3}{2.5cm}{210; Q=116, U=94} &  [Retrieve] & 36\% \\
         & needed or to use & & [No Retrieve] &  21\%    \\
         &  conv. history&  & [Continue to Use Evidence] &   43\% \\  \hline
    
     \multirow{2}{1.5cm}{\texttt{Relevance}}  &  of retrieved passages & \multirow{2}{2.5cm}{301; Q=213, U=88} & [Relevant] &  41\%  \\
      & given conv.  history& &   [Non Relevant] &  59\%  \\    
  \hline 
   \multirow{3}{1.5cm}{\texttt{Groundness}} & generated response & \multirow{3}{2.5cm}{259; Q=229, U=30}  &  [Fully supported] &  43\%  \\
         & is supported by & & [Partially supported] & 13\%   \\      
         & retrieved passage &  & [No support] &  44\%   \\  \hline    
     \multirow{5}{1.5cm}{\texttt{Utility}}  &   & \multirow{5}{2.5cm}{269; Q=143, U=126}  & [Utility:1] & 34\%   \\
       &  usefulness of & &  [Utility:2] &    6\% \\      
        & generated & &  [Utility:3] &    9\% \\
          & response & &[Utility:4] &  23\%\\ 
         & &  &[Utility:5] &   29\%  \\    \bottomrule
    \end{tabular}}
    \caption{Critic training data (QU-MTC) statistics. Q= and U= denotes number of instances sampled from QReCC and UltraChat, respectively.}
    \label{tab:critic_data}
    \vspace{-2mm}
\end{table}

\paragraph{Generator.} The first task of the generator is to generate responses with the special reflection tokens. Given a conversation history as input $x$, we augment the response $y$  to create $\hat{y}$ by including the reflection tokens that is generated by our trained critic model. The generator is trained to generate $\hat{y}$ given $x$ using next token prediction objective. The second task of the generator is to summarise a conversational context to extract important aspects of the conversation \textit{and} pose a question. Given a conversation history, when \selfragmult{} decides retrieval is necessary, it is further prompted to create a summary of the conversation which can be used as query to any retrieval model to obtain passages. We \textit{do not} create a separate critic task for summarization. Rather, we \textit{only} train the generator model since the summarization task is performed \textit{whenever} retrieval is deemed necessary. 

\paragraph{Retriever.} The retriever is the third component of \selfragmult{} and can be used as a separate black-box component. In particular, we use 54M passages from Wikipedia and Common Crawl as the knowledge base.\footnote{The corpus is released as part of the QReCC dataset.} We use an off-the-shelf Contriever model trained on MS-MARCO as the retriever.\footnote{\url{https://github.com/facebookresearch/contriever/}}. The retrieved passages are used by \selfragmult{} to generate responses during inference when it adaptively decides to call retrieval.

\subsection{Overall Framework}
Given a conversation history (that may also include passages retrieved in previous turns), the LLM (generator model) first decides whether retrieval is needed or not by generating one of the three special retrieval tokens: [Retrieve], [No Retrieve], [Continue to use evidence] based on the conversation history and previously retrieved passages. [Retrieve] is typically generated when the response needs to retrieve new facts to respond to a factual question. [No Retrieve] is generated when the question in the conversation requires the model to answer with a creative response. [Continue to use evidence] is typically generated when that the facts needed to answer the current factual question is already present in conversation history or the previously retrieved passages. So no new retrieval is needed and it can rely on the context to generate the response. 

If retrieval is needed, \selfragmult{} then rewrites the conversational history which will be used as a query to retrieve passages from a corpus. The retriever retrieves $K$ passages which the generator process in parallel and retrieves $K$ different candidate outputs conditioned on the conversation history and the retrieved passages. \selfragmult{} then indicates the relevance of each passage to the conversation history by generating the special relevance tokens [Relevant] or [Non Relevant]. Following which \selfragmult{} judges whether the generated responses are [Fully supported], [Partially supported] or [No support] by the respective retrieved passages by generating the corresponding groundedness tokens. Finally, \selfragmult{} gives a usefulness score of [1-5] to the generated response using the utility special token.

The final response is selected out of the candidate responses, using the one which has the highest score in terms of its usefulness, groundedness and the relevance of the passage from which it was generated. Following~\citet{asai2023self}, we conduct a segment-level beam search (with the beam size=$B$) to obtain the top-$B$ segment continuations and return the best sequence at the end of generation. The score of each sequence $y$ with respect to passage $d$ is updated with a score $S$ that is the linear weighted sum of the normalized probability of each special token type:

\begin{align*}
    S = p (y_{t} | x, d, y_{t-1}) & + w_1*S(\texttt{Relevance}) \\
   & + w_2*S(\texttt{Groundedness}) \\
     & + w_3*S(\texttt{Utility})
\end{align*}
where $x$ is the conversation history, $y_{t-1}$ is the generated response so far and $w_{i}$ are hyperparameters that can be tuned to enable custom behavior during inference.\footnote{We use default values defined by~\citet{asai2023self}.} $S(.)$ indicates the generation probability of the most desirable reflection token, e.g., [Relevant] in case of \texttt{Relevance} or [Fully Supported] in case of \texttt{Groundedness}.

 \selfragmult{} is thus capable of comprehending longer conversational contexts (for the various critic tasks) than \selfrag{} and also summarizing the conversational context that when used as a query can improve retrieval effectiveness. The framework is depicted
 in Figure~\ref{fig:selfmultrag}.




\section{Experimental Setup}

 We evaluate \selfragmult{} on three benchmarks: QReCC~\cite{anantha2020open}, UltraChat~\cite{ding2023enhancing} and MT-Eval~\cite{kwan2024mt}. 

 \paragraph{QReCC} contains conversational questions answers to which can be found within a collection of 10M web pages. Answers to questions in the same conversation may be distributed across several web pages. QReCC provides gold passage annotations which indicates the passage from where a question in a conversation can be answered from.
 \paragraph{UltraChat} is traditionally used as supervised fine tuning (SFT) data for LLMs. It contains diverse and informative instructional conversations and covers a wide range of topics and instructions. While conversations in QReCC are knowledge-grounded and ideally RAG should be beneficial for every turn of the conversation, it is not so for UltraChat.

 \paragraph{MT-Eval} is similar to UltraChat as it also contains diverse instructional conversations. It forms our out-of-domain evaluation benchmark as we use samples freom QReCC and UltraChat for training \selfragmult{}. Additional details on the benchmarks are provided in  Appendix~\ref{sec:bench_stat}.

  \begin{table*}[!h]
\small
    \centering
    \begin{tabular}{p{2cm}p{1.5cm}p{1.5cm}p{1.5cm}p{1.5cm}p{1.5cm}p{2cm}p{1cm}}
    \toprule
   \textbf{Dataset} &  \textbf{Model}    & \textbf{\texttt{Ret.} w/o P.} & \textbf{\texttt{Ret.} w P.} &\textbf{\texttt{3-way Ret.}}  & \textbf{\texttt{Relevance}} & \textbf{\texttt{Groundedness}} & \textbf{\texttt{Utility}}   \\ \midrule
&      Critic$_{o}$    & 0.58 & - &0.52 & 0.66 & 0.58 &  0.69  \\
STC  &  Critic$_{s}$  & 0.47& -  & 0.53 & 0.43 & 0.58 & 0.62  \\ 
    single-turn    &  Critic$_{m}$  & 0.50 & - & 0.52 & 0.62 & 0.40 & 0.68 \\ 
            &  Critic$_{sm}$  & \textbf{0.84}& -  & \textbf{0.69}  &  \textbf{0.78} & \textbf{0.71} &\textbf{0.75} \\ \hline
        &      Critic$_{o}$    & 0.43  & 0.51 & 0.28 &  0.61 & 0.35 &  0.46  \\
   QU-MTC:     &  Critic$_{s}$  & 0.68 & 0.28 & 0.62 &0.72 & 0.55 & 0.45 \\ 
        multi-turn   & Critic$_{m}$  & 0.76 & 0.58 & 0.48 & 0.75 & 0.58 & 0.79 \\
               &  Critic$_{sm}$  & \textbf{0.83} & \textbf{0.77} &\textbf{0.63} & \textbf{0.77} & \textbf{0.61} &\textbf{0.80} \\ \bottomrule
    \end{tabular}
    \caption{Critic performance on self-reflection tasks on the test splits of each dataset. We report classification accuracy of predicting the correct special tokens for each task as outlined in Table~\ref{tab:critic_data}. For the \texttt{Retrieval} task, we evaluate critic accuracy, either with passages in the conversation history (\texttt{Ret.} w P.) or without (\texttt{Ret.} w/o P.).}
    \label{tab:critic_acc}
\end{table*}


\subsection{Training Data}


The training data for our critic and generator models are sampled from QReCC and UltraChat.  We employ GPT-4 to collect the labels for training data for critic model. The prompt for collecting GPT-4 labels are provided in the Appendix~\ref{sec:appendix-prompt}. We denote this training data as QReCC-UltraChat Multi-turn Critic Data (QU-MTC), details of which are provided in Table~\ref{tab:critic_data}.
Furthermore, we also create a single-turn variant (QU-STC) by flattening the conversation history to a single-turn using a T5 based query rewriter T5QR~\cite{lin2020conversational}. Thus QU-MTC and QU-STC come from the same data distribution, with the difference being the representation of the conversation history. Lastly, we also use the \textit{original} single-turn critic training dataset, STC, released by~\citet{asai2023self}. 

We employ our trained critic models to create training data for the generator using samples from QReCC and UltraChat (different from those sampled for training the critic), referred to as QReCC UltraChat Multi-turn Generator Data, QU-MTG. As in the case of the critic, we create a single-turn variant of the data, QU-STG, by rewriting the conversation history to a single-turn. Lastly, we sample data from both datasets to create the Conversation Summarization Data (CSD) where we prompt GPT-4 to generate ground truth summaries of $\sim$5000 conversations.\footnote{The prompt for GPT-4 summary collection is provided in Appendix~\ref{sec:appendix-prompt}.} 
\subsection{Models}

We train our own critic models from scratch using \texttt{mistralai/Mistral-7B-Instruct-v0.2} as initial checkpoint. The different versions of critic models are outlined in Table~\ref{tab:model_ver}. Critic$_{o}$ is equivalent to the original critic model trained by~\citet{asai2023self} that is sampled from a number of single-turn benchmarks. However, to understand the impact of training in the conversational setting, Critic$_{sm}$ must be compared with Critic$_{s}$ as they are trained on the same data distribution.

          

   \begin{table}\small
     \centering
     \begin{tabular}{lcll}
     \toprule
         \textbf{Model}  &   \textbf{Critic}& \textbf{Training Data}\\ \midrule
Critic$_{o}$&  - & STC \\
Critic$_{s}$&  - & STC +  QU-STC\\
    Critic$_{m}$      & - &  QU-MTC \\
    Critic$_{sm}$      & - & STC + QU-MTC \\ \midrule
        \selfragorg{} &  Critic$_{s}$ & QU-STG  \\
          \texttt{SELF-RAG$_{sm}$}    &  Critic$_{sm}$ & QU-MTG  \\ 
          
          \selfragmult{}&  Critic$_{sm}$ & QU-MTG + CSD \\\bottomrule
     \end{tabular}
     \caption{Configurations explored as training data for our approaches; ST = single-turn; MT = multi-turn; C = Critic, G = Generator. CSD refers to the conversation summarization dataset.}
     \label{tab:model_ver}
     \vspace{-2mm}
 \end{table}

We also train our generator model from the \texttt{mistralai/Mistral-7B-Instruct-v0.2} checkpoint. The different versions of generator models, that we use to compare the performance of \selfragmult{} with, are also outlined in Table~\ref{tab:model_ver}. 
Henceforth, we use \selfragmult{} to refer to the final model that is trained end-to-end on single-turn, multi-turn conversation data and also trained to summarise conversational context and compare its performance against the other generator models.



\begin{figure*}[!h]
 \centering
 
 \includegraphics[width=\linewidth]{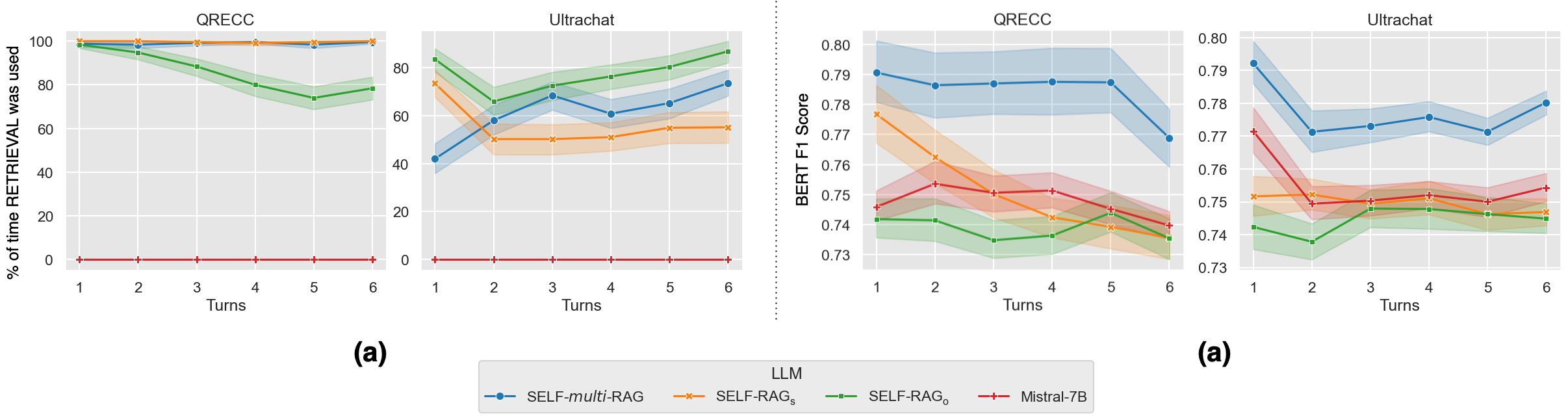}
  \caption{(a) Relation between retrieval calls and number of turns considered in QReCC and UltraChat. (b) Answer quality measured by BERTScore for different turn configurations.}
  \label{fig:QU_plot}
\end{figure*}

\begin{table*}[!h]
    \small
    \centering
    \begin{tabular}{c|ccccccc}
    \toprule
         \textbf{Dataset}  &  \textbf{Model} &  \textbf{GPT4} & \textbf{BERT}   & \textbf{Coherence} & \textbf{Groundness$^*$} & \textbf{Human}\\ \midrule
   \multirow{4}{*}{QReCC}     & \multirow{1}{*}{\texttt{Mistral-7B-Instruct-v0.2}$^-$}   & 2.72 & 0.75  & \textbf{0.98} & - & 4.08\\ 
     & \multirow{1}{*}{\selfrag{}$^\dagger$}  & 1.60 & 0.74 & 0.90 & 0.77 &3.50 \\ 
&   \multirow{1}{*}{\selfragorg{}$^\dagger$} & 1.51 &0.75 & 0.96 &0.78& 3.71 \\
       &  \multirow{1}{*}{\selfragmult{}$^\dagger$} 
& \textbf{4.15} & \textbf{0.80} & 0.97  & \textbf{0.87} & \textbf{4.20} \\\midrule
       \multirow{4}{*}{UltraChat}     & \multirow{1}{*}{\texttt{Mistral-7B-Instruct-v0.2}$^-$}  & 1.71 & 0.75 & 0.99 &- & 3.95 \\ 
     &  \multirow{1}{*}{\selfrag{}$^\dagger$} & 1.57 & 0.74 & 0.95 & 0.89 & 4.00 \\ 
     &   \multirow{1}{*}{\selfragorg{}$^\dagger$} & 1.45 & 0.75 & 0.98& 0.95 &4.05\\
     &   \multirow{1}{*}{\selfragmult{}$^\dagger$}  & 2.70 & \textbf{0.78} & 0.99 & \textbf{0.99 }&\textbf{ 4.46} \\
     \midrule
     \multirow{4}{*}{MT-Eval}     & \multirow{1}{*}{\texttt{Mistral-7B-Instruct-v0.2}$^-$}& 2.04 & 0.62 & 0.92  & - &3.49\\ 
     & \multirow{1}{*}{\selfrag{}$^\dagger$} & 1.50  & 0.71 & 0.94 & 0.82 &3.84\\ 
     &   \multirow{1}{*}{\selfragorg{}$^\dagger$} & 1.20 & 0.72 & 0.95 &0.80 &4.01\\ 
     &   \multirow{1}{*}{\selfragmult{}$^\dagger$}& \textbf{ 3.12 } & \textbf{0.74 }& \textbf{0.96} & \textbf{0.92} &\textbf{4.23}\\ \bottomrule

    \end{tabular}
    \caption{Performance of response generation models on the three dataset. QR = rewritten conversation history as context; FC = full conversation history as context. $^\dagger$ indicates adaptive retrieval; $^-$ indicates no retrieval. $^*$ We measure groundness with retrieved documents only for the cases when the model decides to call retrieval.}
    \label{tab:main_results}
\end{table*}

\subsection{Evaluation}
We first evaluate critic performance on the self-reflection tasks by calculating the accuracy of the predicted tokens described in Table~\ref{tab:critic_data}. We use held out test split of STC and QU-MTC for evaluating critic. To measure the quality of the responses of the generator models, we employ both automatic metrics and human annotations. We collect GPT-4 and human evaluation scores that rate the responses generated by the models on a scale of 1-5 that includes different dimensions such coherence, understandability, and overall quality. The prompt for GPT-4 evaluation is provided in Appendix~\ref{sec:app_GPT4_eval}. We employ BERTScore~\cite{zhang2019bertscore} to measure similarity of generated response with ground truth response. To measure coherence of response given the conversation history and grounding of response given retrieved documents, we employ UniEval~\cite{zhong2022towards}. 

We employ Amazon Mechanical Turk for human evaluation of generated responses. Crowd-workers annotate the quality of the generated responses on a scale of 1 to 5 on the dimensions of coherence, engagingness, and understandability. We collect scores from 3 annotators and aggregate the score using majority voting for each of those dimensions. The overall score is the average of the scores across the three dimensions. In order to ensure high-quality human judgment, we use several mitigation strategies such as simplified task setups, clear annotation guidelines, and time checks to exclude potential spammers. Further details of the human annotation guidelines and disaggregated scores are provided in Appendix~\ref{sec:app_mturk}.


\section{Results}
\label{sec:results}

\paragraph{Critic Performance.} Table~\ref{tab:critic_acc} shows that the critic model trained on both single and multi-turn data Critic$_{sm}$ has overall the best accuracy (generating \textit{correct} reflection tokens) on the critic tasks (based on GPT 4 labels), even improving in the single-turn setting. This suggests that Critic$_{sm}$ is better at handling longer context of conversations while judging whether retrieval is needed or not, judging relevance of retrieved passages, and utility of an answer. 

\paragraph{Response Generation.} As shown in Table~\ref{tab:main_results}, \selfragmult{}, trained on data created by Critic$_{sm}$, leads to improvement on all conversational benchmarks according to the evaluation metrics. Comparing its performance with \selfragorg{}, that has been trained on data from the same dataset but using single-turn contexts, \selfragmult{} performs better in comprehending the conversational context. The decision of whether retrieval is required is more accurate for \selfragmult{} as compared to the baselines. As evidence, we see Critic$_{sm}$ has higher accuracy on retrieval tasks (both with and without passages included in the conversation history) than Critic$_{s}$. Figure~\ref{fig:QU_plot} shows that \selfragmult{} decides to call retrieval $\sim100$\% of time for QReCC, however not so for UltraChat. This is the expected behaviour as conversations in QReCC are mostly knowledge grounded, whereas in UltraChat there are more instructional conversations that not always require retrieved knowledge.  This suggests that the decision to call retrieval or not is indeed important for conversational QA and adapting the model to better handle conversational context is beneficial. Figure~\ref{fig:QU_plot} further shows that \selfragmult{} generated responses are better at all turns (upto 6) of the conversations further providing evidence to its ability to understand long conversational context. Some of the cases where \selfragmult{} called retrieval and could not provide a satisfactory answer is typically the cases where it could not find relevant answers within the retrieved documents. Groundedness of the generated response to the conversational context and retrieved passages, as measured using UniEval, is also higher for \selfragmult{} as compared to its single-turn counterparts. Lastly, we see \selfragmult{} perform the best on MT-Eval, indicating strong performance on held-out conversational benchmarks.

\subsection{Summarizing Conversations}
Since QReCC provides ground truth labels of relevant passages (to a conversational context), we use it to evaluate retrieval effectiveness of different representations of the conversational context. Table~\ref{tab:ret_eff} shows that summaries generated by our approach perform better than rewrites in the form of single-questions (using T5QR) in case of both sparse and dense retrievals. This is in line with research that show expanding queries and documents help in improving retrieval effectiveness~\cite{ayoub2024case,mackie2023generative,nogueira2019doc2query}.

To better understand observed superior performance of \selfragmult{} as compared to its other variants, we perform ablations to narrow down the causes of gain. Table~\ref{tab:diff_summ} reports the performance of different conversation history representations as query to retrieve relevant passages. Note that when we use the \selfragmult{} generated summary as a query to the retrieval model, the response generation is the best for both datasets. Other forms of conversation context as query representations (e.g., T5QR) have lower performance. Overall, \selfragmult{} improves in two directions, (i) it generates summaries as query with better retrieval effectiveness, and (ii) enhances response generation quality taking into account more suitable retrieval knowledge and conversational context.

\begin{table}[!t]
\small
    \centering
    \begin{tabular}{cccc}
    \toprule
      \textbf{Retrieval} & \textbf{Conv. representation}   & \textbf{R@5}  & \textbf{R@10} \\ \midrule
    \multirow{4}{*}{BM25}   &  Full conv. & 0.50 & 0.58 \\
         & Gold Rewrite & 0.50 & 0.61 \\
         & T5QR Rewrite & 0.45 & 0.55 \\
         & \selfragmult{} Summary & 0.56 & 0.66 \\
         \cmidrule{2-4}
         & GPT4 Summary & \textbf{0.60} & \textbf{0.70} \\
          \midrule
          \multirow{4}{*}{Contriever}   &  Full conv. & 0.53 & 0.61 \\
            & Gold Rewrite & 0.58 & 0.69 \\
             & T5QR Rewrite & 0.53 & 0.64 \\
             & \selfragmult{} Summary & 0.61 & 0.71 \\
             \cmidrule{2-4}
             & GPT4 Summary & \textbf{0.62} & \textbf{0.72} \\
             
         \bottomrule
    \end{tabular}
    \caption{Retrieval effectiveness of different conversational context representation. GPT-4 summaries are the ground truth summaries that we collect for training \selfragmult{} for the summarization task.}
    \label{tab:ret_eff}
    \vspace{-2mm}
\end{table}


\subsection{Handling Previous-turn Retrieved Passages}
Returning to Table~\ref{tab:critic_acc}, we also explore the value of previously retrieved passages into the context history of the input for the critic model in the \texttt{Retrieval} critic task. We evaluate critic accuracy, either with previous retrieved passages in the conversation history (\texttt{Ret.} w P.) or without (\texttt{Ret.} w/o P.). For QReCC, we include ground truth passages of previous turn in the context. For UltraChat, we sample instances where passages are present as part of a question in a conversation. As shown in Table~\ref{tab:critic_acc}, Critic$_{sm}$ performs the best in \texttt{Ret.} w P., where the model has to judge whether retrieval is needed or not given both the conversation history and passages retrieved in previous turn. This indicates superior ability of Critic$_{sm}$ to comprehend not only the conversation history but also previously retrieved passages to deem the necessity of retrieval in a multi-turn setting. Overall, the critic models have lower performance in this task compared to the \textit{Retrieval without passages} (\texttt{Ret.} w/o P.) indicating the increased difficulty of the task.

Table~\ref{tab:responsepassage} compares response generation performance with and without retrieved passages from previous turns are included in the input together with the conversation history. We observe that \selfragmult{} not only (correctly) decides to call retrieval less number of times, as compared to its single-turn baseline, but is also better at generating responses when the conversation context is composed of both the dialogue and previously retrieved passages indicating its ability to comprehend more complex contexts. Moreover, either configuration (\texttt{Ret.} w P., \texttt{Ret.} w/o P.) can be chosen based on the desired balance between efficiency and accuracy; \texttt{Ret.} w/o P. is useful when performance is prioritized over efficiency, whereas \texttt{Ret.} w/o P. is suitable when efficiency is crucial and a slight reduction in performance is acceptable. We present examples when passages are included in the context in the Appendix~\ref{sec:examples_resp}.


\begin{table}[!t]
\small
    \centering
    \begin{tabular}{cccc}
    \toprule
      \textbf{Dataset}   & \textbf{Conv. Representation} & \textbf{GPT4} & \textbf{BERT}\\ 
      \midrule
  \multirow{3}{*}{QReCC}       &  Full &  3.75 &  0.786 \\
         & T5QR  &  3.52 & 0.782\\
         &  \selfragmult{} summ. & \textbf{4.15} & \textbf{0.808} \\ \midrule
  \multirow{3}{*}{UltraChat}       & Full &  2.23 &   0.756 \\
         & T5QR & 2.12                     & 0.751 \\
         &  \selfragmult{} summ.& \textbf{2.70} & \textbf{0.780} \\ \midrule
          \multirow{3}{*}{MT-Eval}       & Full &  2.65 &   0.720 \\
         & T5QR & 2.52 &	0.712   \\
         &  \selfragmult{} summ.& \textbf{3.12} & \textbf{0.743} \\ \bottomrule
    \end{tabular}
    \caption{Response generation quality of \selfragmult{} with different conversation history representations as query to retrieve relevant passages.}
    \label{tab:diff_summ}
\end{table}

\section{Conclusion}
\label{sec:conclusion}


In this work, we propose \selfragmult{}, a framework to train LLMs to learn when to retrieve and generate response for better conversational QA. We perform extensive evaluation on three conversational QA benchmarks and demonstrated improved performance over previous approaches. This is achieved by overcoming the previous limitations to accurately critic when to retrieve or whether the retrieved documents are relevant or the usefulness of generated response given a multi-turn dialogue. \selfragmult{} is better at comprehending the longer contexts of multi-turn conversations resulting in better critic and consequently generator performances. Finally, we observe that summaries of conversational history generated by \selfragmult{}, increase retrieval effectiveness when used as query to retrieve passages and consequently leads to improved response generation. As future work, we would like to (i) consider longer and multi-threaded conversations, (ii) including diversity as a metric while considering passages for response generation.

\begin{table}[t]
\small
    \centering
    \resizebox{\linewidth}{!}{
    \begin{tabular}{ccccc}
      \toprule
         \textbf{Dataset}  &  \textbf{Model} & \textbf{Ret.?}
 & \textbf{Groundness}& \textbf{BERT}   \\ \midrule
       \multirow{4}{*}{QReCC}  &\selfragorg{} w/o P.  & 99\% & 0.78 & 0.75\\
       
       & \selfragorg{} w P. &  99\% &  0.73  &  0.74 \\
         &  \selfragmult{} w/o P.& 98\% & 0.87 & 0.80\\
         &  \selfragmult{} w P. & \textbf{ 54\%} &   0.79 & 0.78  \\ \midrule
\multirow{4}{*}{UltraChat}   & \selfragorg{} w/o P. & 56\%  & 0.95 & 0.75 \\

& \selfragorg{} w P. &  67\% &   0.91 & 0.73 \\
    &  \selfragmult{} w/o P.& 62\% & 0.99 & 0.78 \\
         &  \selfragmult{} w P.  &  \textbf{37\%} &  0.97  & 0.75 \\ 
         \midrule
         \multirow{4}{*}{MT-Eval}   & \selfragorg{} w/o P. & 60\%  & 0.80 & 0.72 \\

& \selfragorg{} w P. &  69\% &   0.78 & 0.70 \\
    &  \selfragmult{} w/o P.& 58\% & 0.92 & 0.74 \\
         &  \selfragmult{} w P.  &  \textbf{40\%} &  0.90  & 0.73 \\  \bottomrule
        \end{tabular}}
    \caption{Comparison of response generation performance with and without passages included in the context. Ret? = \% of instances retrieval was used.}
    \label{tab:responsepassage}
    \vspace{-2mm}
\end{table}

\section{Limitations}
\label{sec:limitation}

In this paper, we propose an approach to enhance the ability of retrieved augmented models on conversational settings. While this is not specific to any particular language, we conducted all of our experiments and analysis exclusively on English-language QA datasets. Hence, this paper does not offer insights into the range of style variations found in non-English and datasets, nor does it ascertain the generalizability of our findings to other datasets and domains. Second, we limit our experiments using one model for critic and generator, \texttt{mistralai/Mistral-7B-Instruct-v0.2} and one retrieval model, Contriever. Extending \selfragmult{} to other models is left for future work. Finally, we perform retrieval in an offline manner for reducing computation overhead. In a more realistic scenario, retrieval will be performed online during response generation. 



\bibliography{anthology,custom}

\begin{thebibliography}{38}
\expandafter\ifx\csname natexlab\endcsname\relax\def\natexlab#1{#1}\fi

\bibitem[{Adlakha et~al.(2022)Adlakha, Dhuliawala, Suleman, de~Vries, and Reddy}]{adlakha-etal-2022-topiocqa}
Vaibhav Adlakha, Shehzaad Dhuliawala, Kaheer Suleman, Harm de~Vries, and Siva Reddy. 2022.
\newblock \href {https://doi.org/10.1162/tacl_a_00471} {{T}opi{OCQA}: Open-domain conversational question answering with topic switching}.
\newblock \emph{Transactions of the Association for Computational Linguistics}, 10:468--483.

\bibitem[{Aliannejadi et~al.(2020)Aliannejadi, Chakraborty, R{\'\i}ssola, and Crestani}]{aliannejadi2020harnessing}
Mohammad Aliannejadi, Manajit Chakraborty, Esteban~Andr{\'e}s R{\'\i}ssola, and Fabio Crestani. 2020.
\newblock Harnessing evolution of multi-turn conversations for effective answer retrieval.
\newblock In \emph{Proceedings of the 2020 conference on human information interaction and retrieval}, pages 33--42.

\bibitem[{Anantha et~al.(2020)Anantha, Vakulenko, Tu, Longpre, Pulman, and Chappidi}]{anantha2020open}
Raviteja Anantha, Svitlana Vakulenko, Zhucheng Tu, Shayne Longpre, Stephen Pulman, and Srinivas Chappidi. 2020.
\newblock Open-domain question answering goes conversational via question rewriting.
\newblock \emph{arXiv preprint arXiv:2010.04898}.

\bibitem[{Asai et~al.(2023)Asai, Wu, Wang, Sil, and Hajishirzi}]{asai2023self}
Akari Asai, Zeqiu Wu, Yizhong Wang, Avirup Sil, and Hannaneh Hajishirzi. 2023.
\newblock Self-rag: Learning to retrieve, generate, and critique through self-reflection.
\newblock \emph{arXiv preprint arXiv:2310.11511}.

\bibitem[{Ayoub et~al.(2024)Ayoub, Su, and Li}]{ayoub2024case}
Michael Antonios~Kruse Ayoub, Zhan Su, and Qiuchi Li. 2024.
\newblock A case study of enhancing sparse retrieval using llms.
\newblock In \emph{Companion Proceedings of the ACM on Web Conference 2024}, pages 1609--1615.

\bibitem[{Bai et~al.(2024)Bai, Liu, Bu, He, Liu, Zhou, Lin, Su, Ge, Zheng, and Ouyang}]{bai2024mtbench101}
Ge~Bai, Jie Liu, Xingyuan Bu, Yancheng He, Jiaheng Liu, Zhanhui Zhou, Zhuoran Lin, Wenbo Su, Tiezheng Ge, Bo~Zheng, and Wanli Ouyang. 2024.
\newblock \href {http://arxiv.org/abs/2402.14762} {Mt-bench-101: A fine-grained benchmark for evaluating large language models in multi-turn dialogues}.

\bibitem[{Chiang et~al.(2023)Chiang, Li, Lin, Sheng, Wu, Zhang, Zheng, Zhuang, Zhuang, Gonzalez et~al.}]{chiang2023vicuna}
Wei-Lin Chiang, Zhuohan Li, Zi~Lin, Ying Sheng, Zhanghao Wu, Hao Zhang, Lianmin Zheng, Siyuan Zhuang, Yonghao Zhuang, Joseph~E Gonzalez, et~al. 2023.
\newblock Vicuna: An open-source chatbot impressing gpt-4 with 90\%* chatgpt quality.
\newblock \emph{See https://vicuna. lmsys. org (accessed 14 April 2023)}.

\bibitem[{Culpepper et~al.(2018)Culpepper, Diaz, and Smucker}]{culpepper2018research}
J~Shane Culpepper, Fernando Diaz, and Mark~D Smucker. 2018.
\newblock Research frontiers in information retrieval: Report from the third strategic workshop on information retrieval in lorne (swirl 2018).
\newblock In \emph{ACM SIGIR Forum}, volume~52, pages 34--90. ACM New York, NY, USA.

\bibitem[{Ding et~al.(2023)Ding, Chen, Xu, Qin, Zheng, Hu, Liu, Sun, and Zhou}]{ding2023enhancing}
Ning Ding, Yulin Chen, Bokai Xu, Yujia Qin, Zhi Zheng, Shengding Hu, Zhiyuan Liu, Maosong Sun, and Bowen Zhou. 2023.
\newblock Enhancing chat language models by scaling high-quality instructional conversations.
\newblock \emph{arXiv preprint arXiv:2305.14233}.

\bibitem[{Gao et~al.(2023)Gao, Yen, Yu, and Chen}]{gao2023enabling}
Tianyu Gao, Howard Yen, Jiatong Yu, and Danqi Chen. 2023.
\newblock Enabling large language models to generate text with citations.
\newblock \emph{arXiv preprint arXiv:2305.14627}.

\bibitem[{Guu et~al.(2020)Guu, Lee, Tung, Pasupat, and Chang}]{guu2020retrieval}
Kelvin Guu, Kenton Lee, Zora Tung, Panupong Pasupat, and Mingwei Chang. 2020.
\newblock Retrieval augmented language model pre-training.
\newblock In \emph{International conference on machine learning}, pages 3929--3938. PMLR.

\bibitem[{Ishii et~al.(2022)Ishii, Wilie, Xu, Cahyawijaya, and Fung}]{ishii2022integrating}
Etsuko Ishii, Bryan Wilie, Yan Xu, Samuel Cahyawijaya, and Pascale Fung. 2022.
\newblock Integrating question rewrites in conversational question answering: A reinforcement learning approach.
\newblock In \emph{Proceedings of the 60th Annual Meeting of the Association for Computational Linguistics: Student Research Workshop}, pages 55--66.

\bibitem[{Jang et~al.(2024)Jang, Lee, Bae, Lee, and Jung}]{jang2024itercqr}
Yunah Jang, Kang-il Lee, Hyunkyung Bae, Hwanhee Lee, and Kyomin Jung. 2024.
\newblock Itercqr: Iterative conversational query reformulation with retrieval guidance.
\newblock In \emph{Proceedings of the 2024 Conference of the North American Chapter of the Association for Computational Linguistics: Human Language Technologies (Volume 1: Long Papers)}, pages 8114--8131.

\bibitem[{Ji et~al.(2023)Ji, Gong, Deng, Peng, Niu, Ma, and Li}]{ji2023towards}
Yunjie Ji, Yan Gong, Yong Deng, Yiping Peng, Qiang Niu, Baochang Ma, and Xiangang Li. 2023.
\newblock Towards better instruction following language models for chinese: Investigating the impact of training data and evaluation.
\newblock \emph{arXiv preprint arXiv:2304.07854}.

\bibitem[{Kaiser et~al.(2024)Kaiser, Saha~Roy, and Weikum}]{kaiser2024robust}
Magdalena Kaiser, Rishiraj Saha~Roy, and Gerhard Weikum. 2024.
\newblock Robust training for conversational question answering models with reinforced reformulation generation.
\newblock In \emph{Proceedings of the 17th ACM International Conference on Web Search and Data Mining}, pages 322--331.

\bibitem[{Kulkarni et~al.(2024)Kulkarni, Tangarajan, Kim, and Trivedi}]{kulkarni2024reinforcement}
Mandar Kulkarni, Praveen Tangarajan, Kyung Kim, and Anusua Trivedi. 2024.
\newblock Reinforcement learning for optimizing rag for domain chatbots.
\newblock \emph{arXiv preprint arXiv:2401.06800}.

\bibitem[{Kwan et~al.(2024)Kwan, Zeng, Jiang, Wang, Li, Shang, Jiang, Liu, and Wong}]{kwan2024mt}
Wai-Chung Kwan, Xingshan Zeng, Yuxin Jiang, Yufei Wang, Liangyou Li, Lifeng Shang, Xin Jiang, Qun Liu, and Kam-Fai Wong. 2024.
\newblock Mt-eval: A multi-turn capabilities evaluation benchmark for large language models.
\newblock \emph{arXiv preprint arXiv:2401.16745}.

\bibitem[{Lewis et~al.(2020)Lewis, Perez, Piktus, Petroni, Karpukhin, Goyal, K{\"u}ttler, Lewis, Yih, Rockt{\"a}schel et~al.}]{lewis2020retrieval}
Patrick Lewis, Ethan Perez, Aleksandra Piktus, Fabio Petroni, Vladimir Karpukhin, Naman Goyal, Heinrich K{\"u}ttler, Mike Lewis, Wen-tau Yih, Tim Rockt{\"a}schel, et~al. 2020.
\newblock Retrieval-augmented generation for knowledge-intensive nlp tasks.
\newblock \emph{Advances in Neural Information Processing Systems}, 33:9459--9474.

\bibitem[{Lin et~al.(2020)Lin, Yang, Nogueira, Tsai, Wang, and Lin}]{lin2020conversational}
Sheng-Chieh Lin, Jheng-Hong Yang, Rodrigo Nogueira, Ming-Feng Tsai, Chuan-Ju Wang, and Jimmy Lin. 2020.
\newblock Conversational question reformulation via sequence-to-sequence architectures and pretrained language models.
\newblock \emph{arXiv preprint arXiv:2004.01909}.

\bibitem[{Liu et~al.(2023)Liu, Zhang, and Liang}]{liu2023evaluating}
Nelson~F Liu, Tianyi Zhang, and Percy Liang. 2023.
\newblock Evaluating verifiability in generative search engines.
\newblock \emph{arXiv preprint arXiv:2304.09848}.

\bibitem[{Mackie et~al.(2023)Mackie, Chatterjee, and Dalton}]{mackie2023generative}
Iain Mackie, Shubham Chatterjee, and Jeffrey Dalton. 2023.
\newblock Generative relevance feedback with large language models.
\newblock In \emph{Proceedings of the 46th International ACM SIGIR Conference on Research and Development in Information Retrieval}, pages 2026--2031.

\bibitem[{Mallen et~al.(2023)Mallen, Asai, Zhong, Das, Khashabi, and Hajishirzi}]{mallen2023not}
Alex Mallen, Akari Asai, Victor Zhong, Rajarshi Das, Daniel Khashabi, and Hannaneh Hajishirzi. 2023.
\newblock When not to trust language models: Investigating effectiveness of parametric and non-parametric memories.
\newblock In \emph{Proceedings of the 61st Annual Meeting of the Association for Computational Linguistics (Volume 1: Long Papers)}, pages 9802--9822.

\bibitem[{Mao et~al.(2023{\natexlab{a}})Mao, Dou, Chen, Mo, and Qian}]{mao2023large}
Kelong Mao, Zhicheng Dou, Haonan Chen, Fengran Mo, and Hongjin Qian. 2023{\natexlab{a}}.
\newblock Large language models know your contextual search intent: A prompting framework for conversational search.
\newblock \emph{arXiv preprint arXiv:2303.06573}.

\bibitem[{Mao et~al.(2023{\natexlab{b}})Mao, Qian, Mo, Dou, Liu, Cheng, and Cao}]{mao2023learning}
Kelong Mao, Hongjin Qian, Fengran Mo, Zhicheng Dou, Bang Liu, Xiaohua Cheng, and Zhao Cao. 2023{\natexlab{b}}.
\newblock Learning denoised and interpretable session representation for conversational search.
\newblock In \emph{Proceedings of the ACM Web Conference}, pages 3193--3202.

\bibitem[{Mo et~al.(2023)Mo, Mao, Zhu, Wu, Huang, and Nie}]{mo2023convgqr}
Fengran Mo, Kelong Mao, Yutao Zhu, Yihong Wu, Kaiyu Huang, and Jian-Yun Nie. 2023.
\newblock Convgqr: Generative query reformulation for conversational search.
\newblock \emph{arXiv preprint arXiv:2305.15645}.

\bibitem[{Nogueira et~al.(2019)Nogueira, Lin, and Epistemic}]{nogueira2019doc2query}
Rodrigo Nogueira, Jimmy Lin, and AI~Epistemic. 2019.
\newblock From doc2query to doctttttquery.
\newblock \emph{Online preprint}, 6(2).

\bibitem[{Oh and Thorne(2023)}]{oh2023detrimental}
Philhoon Oh and James Thorne. 2023.
\newblock Detrimental contexts in open-domain question answering.
\newblock \emph{arXiv preprint arXiv:2310.18077}.

\bibitem[{Ram et~al.(2023)Ram, Levine, Dalmedigos, Muhlgay, Shashua, Leyton-Brown, and Shoham}]{ram2023context}
Ori Ram, Yoav Levine, Itay Dalmedigos, Dor Muhlgay, Amnon Shashua, Kevin Leyton-Brown, and Yoav Shoham. 2023.
\newblock In-context retrieval-augmented language models.
\newblock \emph{arXiv preprint arXiv:2302.00083}.

\bibitem[{Shi et~al.(2023)Shi, Chen, Misra, Scales, Dohan, Chi, Sch{\"a}rli, and Zhou}]{shi2023large}
Freda Shi, Xinyun Chen, Kanishka Misra, Nathan Scales, David Dohan, Ed~H Chi, Nathanael Sch{\"a}rli, and Denny Zhou. 2023.
\newblock Large language models can be easily distracted by irrelevant context.
\newblock In \emph{International Conference on Machine Learning}, pages 31210--31227. PMLR.

\bibitem[{Wang et~al.(2023)Wang, Song, Xu, and Su}]{wang2023domain}
Ante Wang, Linfeng Song, Ge~Xu, and Jinsong Su. 2023.
\newblock Domain adaptation for conversational query production with the rag model feedback.
\newblock In \emph{Findings of the Association for Computational Linguistics: EMNLP 2023}, pages 9129--9141.

\bibitem[{Wei et~al.(2024)Wei, Yang, Song, Lu, Hu, Huang, Tran, Peng, Liu, Huang, Du, and Le}]{wei2024longform}
Jerry Wei, Chengrun Yang, Xinying Song, Yifeng Lu, Nathan Hu, Jie Huang, Dustin Tran, Daiyi Peng, Ruibo Liu, Da~Huang, Cosmo Du, and Quoc~V. Le. 2024.
\newblock \href {http://arxiv.org/abs/2403.18802} {Long-form factuality in large language models}.

\bibitem[{Wu et~al.(2024)Wu, Wu, and Zou}]{wu2024faithful}
Kevin Wu, Eric Wu, and James Zou. 2024.
\newblock \href {http://arxiv.org/abs/2404.10198} {How faithful are rag models? quantifying the tug-of-war between rag and llms' internal prior}.

\bibitem[{Wu et~al.(2021)Wu, Luan, Rashkin, Reitter, Hajishirzi, Ostendorf, and Tomar}]{wu2021conqrr}
Zeqiu Wu, Yi~Luan, Hannah Rashkin, David Reitter, Hannaneh Hajishirzi, Mari Ostendorf, and Gaurav~Singh Tomar. 2021.
\newblock Conqrr: Conversational query rewriting for retrieval with reinforcement learning.
\newblock \emph{arXiv preprint arXiv:2112.08558}.

\bibitem[{Xu et~al.(2023)Xu, Shi, and Choi}]{xu2023recomp}
Fangyuan Xu, Weijia Shi, and Eunsol Choi. 2023.
\newblock Recomp: Improving retrieval-augmented lms with compression and selective augmentation.
\newblock \emph{arXiv preprint arXiv:2310.04408}.

\bibitem[{Ye et~al.(2023)Ye, Fang, Li, and Yilmaz}]{ye2023enhancing}
Fanghua Ye, Meng Fang, Shenghui Li, and Emine Yilmaz. 2023.
\newblock Enhancing conversational search: Large language model-aided informative query rewriting.
\newblock \emph{arXiv preprint arXiv:2310.09716}.

\bibitem[{Yoran et~al.(2023)Yoran, Wolfson, Ram, and Berant}]{yoran2023making}
Ori Yoran, Tomer Wolfson, Ori Ram, and Jonathan Berant. 2023.
\newblock Making retrieval-augmented language models robust to irrelevant context.
\newblock \emph{arXiv preprint arXiv:2310.01558}.

\bibitem[{Zhang et~al.(2019)Zhang, Kishore, Wu, Weinberger, and Artzi}]{zhang2019bertscore}
Tianyi Zhang, Varsha Kishore, Felix Wu, Kilian~Q Weinberger, and Yoav Artzi. 2019.
\newblock Bertscore: Evaluating text generation with bert.
\newblock \emph{arXiv preprint arXiv:1904.09675}.

\bibitem[{Zhong et~al.(2022)Zhong, Liu, Yin, Mao, Jiao, Liu, Zhu, Ji, and Han}]{zhong2022towards}
Ming Zhong, Yang Liu, Da~Yin, Yuning Mao, Yizhu Jiao, Pengfei Liu, Chenguang Zhu, Heng Ji, and Jiawei Han. 2022.
\newblock Towards a unified multi-dimensional evaluator for text generation.
\newblock \emph{arXiv preprint arXiv:2210.07197}.

\end{thebibliography}
\bibliographystyle{acl_natbib}

\appendix
\section*{Appendix}
\section{GPT-4 Prompt: Critic Training Data \&  Conversation Summarization Training Data}
\label{sec:appendix-prompt}

    Prompts to collect critic training data for the different tasks using GPT-4 are outlined in Table~\ref{tab:critic_ret} (\texttt{Retrieval}), Table~\ref{tab:critic_mult_ret} (\texttt{3-way Retrieval}), Table~\ref{tab:critic_rel} (\texttt{Relevance}), Table~\ref{tab:critic_ground} (\texttt{Groundedness}), Table~\ref{tab:critic_utility} (\texttt{Utility}), We modify the prompts of~\citet{asai2023self} to be compatible with multi-turn dialogues (as compared to single-turn QA). Furthermore, we collect conversation summarization training data for our generator model using GPT-4 prompt detailed in Table~\ref{tab:gen_summary}.

\begin{table*}
    \centering
    \begin{tabular}{p{15cm}}
       \toprule
     \rowcolor[gray]{0.8}  \textbf{\texttt{Retrieval}} \\ \midrule
       Given \textbf{\textit{a conversation history}}, please make a judgment on whether finding some external documents from the web (e.g., Wikipedia) helps to generate a better response. Please answer [Retrieval] or [No Retrieval].  \\ \hdashline
     \textbf{Conversation History}\\
    
    How did the Boer war start? \\
    
    Many historians stress that in reality the contest was for control of the rich Witwatersrand gold-mining complex located in the SAR. \\
    
    What were the Boer Commandos?\\
    
    \textbf{Rating}: [Retrieval] \\
    
    \textbf{Explanation}: Retrieving  documents will help in generating a good response to the conversation.\\
    
    \textbf{Conversation History}\\
    
    How does the taste and texture of swordfish change depending on how it's prepared, and what are some popular cooking methods?\\
    
    The taste and texture of swordfish can vary greatly depending
    on how it is prepared. Here are some popular cooking methods and how they affect the taste and texture of the fish: 1. Grilling: Grilling swordfish gives it a smoky flavor and a slightly charred texture. The meat becomes firmer and less flaky. 2. Broiling: Broiling swordfish can give it a crispy exterior while keeping the inside moist and tender. 3. Baking: Baking swordfish at a high temperature can create a crust on the outside of the fish while keeping the inside moist and tender. 4. Pan-searing: Searing swordfish in a hot skillet with oil can give it a crispy exterior while keeping the inside moist and flaky. 5. Poaching: Poaching swordfish in a liquid such as wine, lemon juice or stock can add flavor to the fish and keep it moist and tender. 6. Raw: Swordfish can also be served raw as sushi or sashimi. Raw swordfish has a firm texture and a mild, sweet flavor. \\
    
    I've only ever had swordfish grilled, what other methods do you recommend trying?\\
    
    \textbf{Rating}: [No Retrieval]\\
    
    \textbf{Explanation}: The response doesn't require any external evidence as it can be generated based on the conversation history.\\
    \\
    \textbf{Conversation History}\\

     when did Joe Satriani start his music career? \\
     
     Satriani started playing in a San Francisco-based band called the Squares what instrument does Joe Satriani play? \\
     
     He was inspired to play guitar at age 14, after hearing of the death of Jimi Hendrix. \\
     
     what year did he start his music career?\\
    
    \textbf{\textcolor{blue}{GPT-4-Rating}}: [Retrieval] \\ \bottomrule
    \end{tabular}
    \caption{GPT-4 prompt for collecting training data for Retrieval critic task.}
    \label{tab:critic_ret}
\end{table*}
    
\begin{table*}
    \centering
    \small
    \begin{tabular}{p{15cm}}
       \toprule
     \rowcolor[gray]{0.8}  \textbf{\texttt{3-way-Retrieval}} \\ \midrule
You will be provided with \textbf{\textit{a conversation history}}, evidence, response to the \textbf{\textit{conversation}}, and preceding sentences (optional). If the preceding sentence is given, the response should be the sentence that follows those preceding sentences. Your task is to determine whether the information in the response sentence can be fully verified by the evidence or the \textbf{\textit{conversation history}}. There are three cases: \\
    
    - If the response can be verified solely with the evidence or the \textbf{\textit{conversation history}}, then respond with [Continue to Use Evidence]. \\
    
    - If the sentence doesn't require any factual verification (e.g., a subjective sentence or a sentence about common sense), then respond with [No Retrieval]. \\
    
    - If additional information is needed to verify the output sentence, respond with [Retrieval]. Please provide explanations for your judgments.\\ \hdashline
     \textbf{Conversation History}\\
    
    Given the text: that truth might finally win. The candidates deceive us all. in jets and campaign buses. The smiles they wore were painted on. So sly, those hungry foxes. on top of their soapboxes. "Hey, I'll do much more taxing. so you can be relaxing." I'll give you fruits of their hard work. "You've heard me, one and all! We need a woman president! We need a leader NOW! We need to have a first "first man" to be since smoking pot in college with Bill.. because I JUST WANT TO BE PRESIDENT!! We do not need a woman. but someone true who can. for answers to be found. so maybe we should look. So why is he not jailed? Now here's another "great" debate. What difference does it really make? It tells me this: Who cares? \\
    
    Can you summarize the text material to describe the main message and theme it conveys? \\
    
    The text criticizes political candidates and their deceitful tactics during campaigns, highlighting their insincerity and lack of concern for the public. The message suggests a need for a truthful and effective leader to bring about change and progress, rather than focusing on superficial qualities such as gender or past indiscretions. The theme centers around the importance of honesty and integrity in leadership\\ 
    
    Can you provide examples from the text that show the candidates' insincerity and deceitful tactics during campaigns? \\
    
    \textbf{Preceding sentences}: Here are a few examples from the text that demonstrate the candidates' insincerity and deceitful tactics during campaigns.\\
    
    \textbf{Evidence}: The charisma of the sender of a message may affect how the message is received. Political candidates are often chosen more for their possession of this quality than for their other attributes. A charismatic person can often make tired, trivial messages seem new and important to the recipient; however, this too can become detrimental to communication, as the receiver of the message is less likely to question or ask for clarification of the message. \\
    
    \textbf{Response}: 'The smiles they wore were painted on' suggests that the candidates are not genuinely happy or friendly, but are instead putting on a façade to deceive the public. - 'So sly, those hungry foxes' implies that the candidates are cunning and opportunistic, willing to say or do whatever it takes to win. - 'Hey, I'll do much more taxing. So you can be relaxing.' This statement is a classic political promise that is often made but rarely kept, highlighting the candidates' tendency to make unrealistic claims in order to garner support. \\
    
    \textbf{Rating}: [Continue to Use Evidence] \\
    
    \textbf{Explanation}: The response can be generated solely using the conversational history.\\ \\
    \textbf{Conversation History}\\
    In a small sauce pot on low heat, melt butter and add the vegetable oil. Add onions and cook on low heat until they are browned and soft. Approximately 20 minutes. Add sugar and cook for 2 minutes. Stir in vinegar and stock and cook for an additional 10 minutes or until gravy has reduced by half. Stir in mustard and pepper. Season to taste with salt. Remove from heat and using a hand blender, blend until smooth. Is there a substitute for vegetable oil that can be used in this recipe?\\
    
    Yes, you can use any other type of neutral oil such as canola oil, grapeseed oil, or sunflower oil as a substitute for vegetable oil in this recipe.\\
    
    Can you give me some tips on how to make sure the onions are properly browned and soft?\\
    
    \textbf{Preceding sentences}: Sure! Here are some tips to ensure your onions are browned and soft:  1. Use a low heat: Cooking the onions on low heat allows them to slowly caramelize and become sweet and tender.  2. Stir regularly: Stirring the onions regularly will help prevent them from burning and ensure that they are evenly browned.  3. Use a heavy-bottomed pot: A heavy-bottomed pot ensures even heat distribution and helps prevent burning.  4. Don't overcrowd the pot: Overcrowding the pot will cause the onions to steam instead of caramelize, so make sure there is enough room for them to spread out. \\
    
    \textbf{Evidence}:  Palm oil adds some great firmness to your soap, and there is no direct replacement for it. Grapeseed and sunflower oil are both liquid at room temperature. They’re considered soft oils, and will make this recipe softer than if it had palm oil. You can leave palm oil out. \\
    
    \textbf{Response}:   5. Be patient: Properly caramelizing onions takes time, so don't rush the process.\\

    \textbf{\textcolor{blue}{GPT-4-Rating}}: [No Retrieval] \\
     \bottomrule
    \end{tabular}
    \caption{GPT-4 prompt for collecting training data for Multi-Retrieval critic task.}
    \label{tab:critic_mult_ret}
\end{table*}

\begin{table*}
    \centering
    \begin{tabular}{p{15cm}}
       \toprule
     \rowcolor[gray]{0.8}  \textbf{\texttt{Relevance}} \\ \midrule
You’ll be provided with \textbf{\textit{a conversation history}}, along with an evidence.Your job is to determine if the evidence is relevant and provides useful information to generate the response of the given \textbf{\textit{conversation history}}. If the evidence meets this requirement, respond with [Relevant]; otherwise, generate [Irrelevant]. \\ \hdashline 
 \textbf{Conversation History}:\\ 
    
    How did the Boer war start? \\ 
    
    Many historians stress that in reality the contest was for control of the rich Witwatersrand gold-mining complex located in the SAR. \\ 
    
    What were the Boer Commandos?\\ 
    
    \textbf{Evidence}: Boer Commando Not to be confused with Commando System (South Africa) or Kommandokorps. The Boer commandos or " Kommandos " were volunteer military units of guerilla militia organized by the Boer people of South Africa . The term came into English usage during the Second Boer War of 1899-1902. Boer Commando in action during the First Boer War , 1881 In 1658, war erupted between the Dutch settlers at Cape Colony and the Khoi-khoi . In order to protect the settlement, all able bodied men were conscripted. After the conclusion of this war, all men in the colony were liable for military service and were expected to be ready on short notice.\\  
    
    \textbf{Rating}: [Relevant] \\ 
    
    \textbf{Explanation}: The evidence explicitly talks about Boer commandos from which the response to the conversation can be generated. \\ 
    
    \textbf{Conversation History}: \\ 
    
    What was the origin of the Olmec? \\ 
    
    The beginnings of Olmec civilization have traditionally been placed between 1400 and 1200 BCE. It seems that the Olmec had their roots in early farming cultures of Tabasco. \\ 
    
    What can you tell me about the Olmec at El Manati? \\ 
    
    Past finds of Olmec remains were ritually deposited at El Manati shrine. \\ 
    
    How did they start?\\ 
    
    \textbf{Evidence}: It is a theory that according to many, could explain the incredible technologies and skills of this enigmatic Ancient Civilization. Even though the Olmec civilization is surrounded by numerous mysteries, researchers believe that all the classical cultures of Mesoamerica originated from this mysterious civilization. But where did this ancient civilization originate? And why is it that we know so little about one of the most influential ancient civilizations of Mesoamerica. \\ 
    
    \textbf{Rating}: [Irrelevant] \\ 
    
    \textbf{Explanation}: Although the evidence talks about Olmecs, they do not provide information as to how the Olmec civilisation started. \\ \\ 
    
    \textbf{Conversation History}\\ 
    
    Where did J. R. Jayewardene live in his early life? \\ 
    
    J. R. Jayewardene was born in Colombo, British Ceylon. \\ 
    
    Where did he go to school? \\ 
    
    \textbf{Evidence}: Highly respected in Japan for his call for peace and reconciliation with post-war Japan at the Peace Conference in San Francisco in 1951, a statue of Jayewardene was erected at the Kamakura Temple in the Kanagawa Prefecture in Japan in his honor. [18] J.R Jayewardene Centre [ edit ] In 1988, the J.R. Jayewardene Centre was established by the J.R Jayewardene Centre Act No. 77 of 1988 by Parliament at the childhood home of J. R. Jayewardene Dharmapala Mawatha, Colombo. It serves as archive for J.R Jayewardene's personal library and papers as well as papers, records from the Presidential Secretariat and gifts he received in his tenure as \\ President. 

    \textbf{\textcolor{blue}{GPT-4-Rating}}: [Irrelevant] \\

          \bottomrule
    \end{tabular}
    \caption{GPT-4 prompt for collecting training data for Relevance critic task.}
    \label{tab:critic_rel}
\end{table*}

\begin{table*}
    \centering
    \begin{tabular}{p{15cm}}
       \toprule
     \rowcolor[gray]{0.8}  \textbf{\texttt{Groundedness}} \\ \midrule

 You will receive\textbf{\textit{ a conversation history}}, evidence, and a response to the \textbf{\textit{conversation}}. Your task is to evaluate if the response is fully supported by the information provided in the evidence or \textbf{\textit{in the conversation history}}. Use the following entailment scale to generate a score:\\ 
    
    [Fully supported] - All information in output is supported by the evidence, or extractions from the evidence or \textbf{the conversation history}. \\
    
    [Partially supported] - The response is supported to some extent, but there is major information in the response that is not discussed in the evidence or the conversation history. \\
    
    [No support / Contradictory] - The response completely ignores, is unrelated to, or contradicts the evidence and the conversation history. This can also happen if the evidence is irrelevant to the conversation history. Make sure to not use any external information/knowledge to judge whether the response is true or not. \\ \hdashline
     \textbf{Conversation History}\\
    
    I was thinking of buying a cheesecake, can you tell me some information about them? \\
    
    Sure! Cheesecakes are actually my speciality. Usually it is a baked dessert but it can also be unbaked.\\ 
    
    I had no idea you didn't have to bake them, is there any interesting history behind the cheesecake? \\
    
    Forms of it go back all the way to greece! \\
    
    That's fascinating, can you tell me more about the Greek cheescake origins? \\
    
    Sure. The earliest mentions of it were in a Greek book. Essentially it was a cookbook about the art of making cheesecakes I wonder why they were so fond of them.\\
    
    Is there a traditional recipe for modern cheesecakes? \\
    
    \textbf{Response}: My favourite layer is the biscuit base, particularly if ginger biscuits are mixed in with the graham crackers. \\
    
    \textbf{Evidence}: Cheesecakes, having a crust that is separately prepared and baked. A more modern version is found in Forme of Cury', an English cookbook from 1390. On this basis, chef Heston Blumenthal has argued that cheesecake is an English invention. Cheesecake did not evolve into the dessert that we see today up until somewhere around the 18th century. Europeans began removing yeast and adding beaten eggs to the cheesecake instead. \\
    
    \textbf{Rating}: [No support / Contradictory]  \\
    
    \textbf{Explanation}: The response is neither supported by the evidence or the conversation history \\ \\
    
   \textbf{Conversation History}\\
   
  What does it mean to be a vegan?\\
  
  Veganism is a stricter form of vegetarianism; like vegetarians, vegans don't eat any animals that have been killed and also avoid animal products altogether, so that means no eggs or dairy products and, usually, no honey either.\\
  
  What is the environmental argument for being a vegan?\\
  
  Studies on world food security estimate that an affluent diet containing meat requires up to 3 times as many resources as a vegetarian diet\\
  
  Do you need milk for strong bones?\\
  
  If you want to drink milk for strong bones, I recommend no more than one glass a day\\
  
  What are other good sources of calcium\\
  
  \textbf{Response}: However, many non-dairy sources are also high in this mineral. These include seafood, leafy greens, legumes, dried fruit, tofu and various foods that are fortified with calcium. \\
  
  \textbf{Evidence}: Top 15 Calcium-Rich Foods (Many Are Non-Dairy) Subscribe Nutrition Evidence Based Top 15 Calcium-Rich Foods (Many Are Non-Dairy) Written by Kerri-Ann Jennings, MS, RD on July 27, 2018 If you buy something through a link on this page, we may earn a small commission. How this works. Calcium is very.\\

  \textbf{\textcolor{blue}{GPT-4-Rating}}: [Partially supported] \\
               \bottomrule
    \end{tabular}
    \caption{GPT-4 prompt for collecting training data for Groundedness critic task.}
    \label{tab:critic_ground}
\end{table*}

\begin{table*}
    \centering \small
    \begin{tabular}{p{15cm}}
       \toprule
     \rowcolor[gray]{0.8}  \textbf{\texttt{Utility}} \\ \midrule
 Given a conversation history and a response, rate whether the response appears to be a helpful and informative answer to the query, from 1 (lowest) - 5 (highest). We call this score perceived utility.The detailed criterion is as follows:\\
    
    5: The response provides a complete, highly detailed, and informative response to the conversation, fully satisfying the information needs.\\
    
    4: The response mostly fulfills the need in the conversation, while there can be some minor improvements such as discussing more detailed information, having better structure of the response, or improving coherence.\\
    
    3: The response is acceptable, but some major additions or improvements are needed to satisfy users’ needs.\\
    
    2: The response still addresses the main request, but it is not complete or not relevant to the conversation.\\
    
    1: The response is barely on-topic or completely irrelevant. \\ \hdashline
     \textbf{Conversation History} \\ 
    
    What was snake river canyon for Evel Knievel? \\ 
    
    Knievel hired subcontractor and aeronautical engineer Doug Malewicki to build him a rocket-powered cycle to jump across the Snake River, and called it the Skycycle X-1.\\ 
    
    What was the Evel Knievel's snake river canyon jump a launch of? \\ 
    
    The decision was then made to have Truax build the Skycycle X-2 and have it take off and fly more like a rocket than a motorcycle. \\ 
    
    When was the launch? \\ 
    
    \textbf{Response}: It was launched on September 7, 1974. \\ 
    
    \textbf{Perceived utility}: 5 \\ 
    
    \textbf{Explanation}: The response is accurate, brief, and directly addresses the user's question. It provides the exact date of the launch, which is the specific information that the user was asking for.\\ 
    
    \textbf{Conversation History:} \\ 
    
    How do elephants use their trunks for communication and sensing their environment? \\ 
    
    Elephants use their trunks for communication and sensing their environment in several ways: 1. Smelling: Elephants have a highly developed sense of smell and can detect scents from miles away. They use their trunks to smell for food, water, mates, predators, and other elephants. They can also smell the scent left by other elephants as a means of identifying them. 2. Touch: Elephants use their trunks to touch and feel their environment. They can use their trunks to brush away leaves or branches to get a clear view of their surroundings, and they can touch other elephants to communicate their mood or intentions. 3. Vocalizations: Elephants can produce a wide range of vocalizations, from low rumbles to high-pitched trumpets. These sounds are produced by muscles in their trunks and are used for communication with other elephants. \\ 
    
    I wonder if elephants have ever been observed using their trunks to communicate with other animals besides other elephants? \\ 
    
    \textbf{Response}: I'm not sure, but elephants are known to be highly intelligent and social animals, so it's possible that they have developed complex communication systems with other animals as well. \\ 
    
    \textbf{Perceived utility}: 2 \\ 
    
    \textbf{Explanation}: The assistant did provide some relevant information about elephants being highly intelligent and social animals, but doesn't fully satisfy the user's need for a definite answer. \\ \\ 
    
    \textbf{Conversation History}: \\ 
    
    What was the relation between Dimebag and Darrell?\\ 
    
    Dimebag Darrell was an American musician and songwriter. \\ 
    
    He was the guitarist of the heavy metal bands Pantera and Damageplan. \\ 
    
    When was the band Pantera formed? \\ 
    
    Pantera was formed in 1981. \\ 
    
    What was the purpose of the band? \\ 
    
    \textbf{Response}: Vinnie Paul was asked to join a band alongside his high school classmates Terry Glaze (guitar), Tommy Bradford (bass) and Donny Hart (vocals). \\ 

  \textbf{\textcolor{blue}{GPT-4-Rating}}: 2
    \\\bottomrule
    \end{tabular}
    \caption{GPT-4 prompt for collecting training data for Utility critic task.}
    \label{tab:critic_utility}
\end{table*}

\begin{table*}
    \centering 
    \begin{tabular}{p{15cm}}
       \toprule
     \rowcolor[gray]{0.8}  \textbf{Conversation summarization} \\ \midrule
Given a conversation history, your task is to summarise the conversation history in 40-50 words and ask a question so that the summary and the question can be used without the conversation history to generate a meaningful response. \\ \hdashline

\textbf{Converation History}: \\
What was the first job John Sherman Cooper held?\\ 
He was admitted to the bar by examination in 1928 and opened a legal practice in Somerset.\\
What was the first office John Sherman Cooper ran for?\\
After being urged into politics by his uncle, Judge Roscoe Tartar, Cooper ran unopposed for a seat in the Kentucky House of Representatives as a Republican in 1927.\\
How long was John Sherman Cooper in office in the Kentucky House of Representatives?\\
Member of the Kentucky House of Representatives from the 41st district. In office, 1928–1930\\
Did he run for another political office after that?\\
\textbf{Summary:}  John Sherman Cooper started his career as a lawyer in Somerset after being admitted to the bar in 1928. He was later encouraged by his uncle, Judge Roscoe Tartar, to join politics and subsequently ran for a seat in the Kentucky House of Representatives as a Republican candidate in 1927. He went unopposed and served in office from 1928-1930. \\

\textbf{Question}: Did John Sherman Cooper pursue any other political offices after his term in the Kentucky House of Representatives? \\ 

\textbf{Converation History}: \\
When did Sachin Tendulkar first join a team?\\
On 14 November 1987, Sachin Tendulkar was selected to represent Bombay in the Ranji Trophy, India's premier domestic First-class cricket tournament, for the 1987–88 season.\\
Was he successful with that team? \\

\textbf{Summary}: Sachin Tendulkar first joined the Bombay team in the Ranji Trophy, India's premier domestic First-class cricket tournament, on November 14, 1987. \\
\textbf{Question}: Was Sachin Tendulkar's performance successful with this team? 
     \\\bottomrule
    \end{tabular}
    \caption{GPT-4 prompt for collecting generator training data for conversation summarization.}
    \label{tab:gen_summary}
\end{table*}

\section{Benchmark Statistics}
 \label{sec:bench_stat}
 While QReCC is traditionally used as the primary benchmark for evaluating conversational query rewriting models or response generation, we note a few differences with UltraChat and MT-Eval. Firstly, as mentioned earlier, responses to all conversation turns are grounded in passage from Wikipedia/Common Crawl. Hence, ideally retrieval should be needed at every turn, and hence RAG should lead to the best response. Whereas in the other two datasets there are examples where a passage is already included in a conversation turn or ones where the response to a turn should be based on response generated in previous turns or conversations where users ask creative questions (examples provided in Table~\ref{tab:data_stats}). Hence, in these two datasets the decision to call retrieval is more important where a model should understand from the context whether retrieval is needed or not. Secondly, questions and gold responses in QReCC are an order of magnitude shorter than UltraChat as shown in Table~\ref{tab:data_stats}). 
\begin{table*}
    \small
    \centering
    \resizebox{\linewidth}{!}{
    \begin{tabular}{p{1.5cm}|p{5cm}|p{5cm}|p{4.7cm}}
     \toprule
      \textbf{}   &  \textbf{QReCC} & \textbf{Ultrachat}  &  \textbf{MT-Eval } \\ \midrule
     Avg. \# words in one question   & 7 &  19 & 32\\ \hline
      Avg. \# words in response    & 67  & 208   & 144\\ \midrule
    Example conversation     &  \textbf{User:} which is the best jekyll and hyde film? &   \textbf{User:} Provide a detailed analysis of the role of artificial intelligence in enhancing customer experience within the hospitality industry. & \textbf{User:} What is the central dogma of molecular biology? What processes are involved? Who named this? \\ & \textbf{Assistant:} The 1931 film adaptation of Robert Louis Stevenson's horror novella is considered to be the best Jekyll an Hyde film of all time.  & \textbf{Assistant:} Artificial intelligence (AI) has been playing an increasingly important role in enhancing customer experience in various industries, including hospitality. Hospitality businesses such as hotels, restaurants, and travel companies are using AI to improve their customer service, streamline processes and gain a competitive edge.  Below are some of the ways AI is transforming customer experience in the hospitality industry:  1. Personalized recommendations and experiences  AI-powered systems can analyze guest data such as booking history, preferences, demographics, and behavior to provide personalized recommendations and experiences. For instance, a hotel can use AI to suggest personalized room upgrades, recommend nearby attractions, or offer personalized dining recommendations based on guests' dietary requirements.  2. Chatbots and virtual assistants  ....    In summary, AI is transforming the hospitality industry by enhancing customer experience, increasing efficiency, and reducing costs. With the implementation of AI, businesses can provide personalized experiences, streamline processes, and gain a competitive edge. As AI technology continues to improve, it is likely to become an essential aspect of the hospitality industry. & \textbf{Assistant:} The central dogma of molecular biology is a fundamental principle that describes the flow of genetic information within a biological system. It was first proposed by Francis Crick in 1958. The central dogma states that genetic information flows from DNA to RNA to proteins, which are responsible for the structure and function of cells. …  In summary, the central dogma of molecular biology, proposed by Francis Crick, describes the flow of genetic information from DNA to RNA to proteins, involving the processes of DNA replication, transcription, and translation.\\ & \textbf{User:} when was strange case of dr jekyll and mr hyde published   & \textbf{User:} I love the idea of personalized recommendations and experiences. Can you give me an example of a hotel that effectively uses AI for this purpose? &  \textbf{User:} Identify and fix one incorrect fact in your previous response.\\ & \textbf{Assistant}: Strange Case of Dr Jekyll and Mr Hyde is a gothic novella by Scottish author Robert Louis Stevenson, first published in 1886. & 
 \textbf{Assistant: }Sure! One example of a hotel that effectively uses AI for personalized recommendations and experiences is the Henn-na Hotel in Japan. This hotel uses robot receptionists and facial recognition technology to provide personalized experiences for their guests.  ...  Overall, the use of AI in Henn-na Hotel provides a seamless and personalized hospitality experience that's entirely automated while maximizing customer service. & \textbf{Assistant:} Apologies for the mistake in my previous response. The central dogma of molecular biology was first proposed by Francis Crick in 1957, not 1958. The rest of the information provided remains accurate.
 \\ &\textbf{User:} what other books did the author write" & \textbf{User:} Wow, that sounds really futuristic. I'd love to stay at that hotel and experience all of their AI-powered amenities. Do you know if more hotels are adopting these technologies? &\textbf{User:} Can you elaborate on how DNA replication in stem cells differs from that in other types of cells, particularly in the context of its accuracy and regulation? \\ \bottomrule
    \end{tabular}}
    \caption{Statistics and examples for the three datasets considered in this study.}
    \label{tab:data_stats}
\end{table*}

\section{Implementation Details}
\label{sec:imp_det}

We use \texttt{Mistral-7B-Instruct-v0.2} as our critic an generator models. Critic training is performed using 4 nodes, each containing 8 NVIDIA A10G Tensor Core GPUs. The models are trained for 3 epochs, learning rate of $2e^{-5}$ and warm up ratio of $0.01$. Generator training is performed using 5 nodes, 2 epochs, learning rate of $2e^{-5}$ and warm up ratio of $0.03$. The distributed training is performed using deepspeed\footnote{\url{https://github.com/microsoft/DeepSpeed}}. For computing BERT-Score, we use \texttt{roberta-large-mnli} as our model.

\section{GPT-4 Evaluation Form}
\label{sec:app_GPT4_eval}
We employ GPT-4 to evaluate the responses generated from the different SELF-RAG models. We set temperature to be $1.0$ and maximum number of generated tokens (\texttt{max\_tokens}) to be $512$. The prompt for GPT-4 evaluation is detailed in Table~\ref{tab:GPT4_eval}.

\begin{table*}
    \centering
 \begin{tabular}{p{\linewidth}}
 \toprule
        \textbf{Prompt for GPT-4 Evaluation} \\ \toprule
     \textbf{Conversation History:} \\
\texttt{conversation} \\
\textbf{The Last User Question:} \\
\texttt{question} \\
\textbf{The Start of Assistant's Answer} \\
\texttt{generated\_response} \\
\textbf{The End of Assistant's Answer} \\
\textbf{System:} \\
Given the conversation history given above, we would like to request your feedback on the performance of the assistant in response to the \textbf{Last User Question} as displayed above. Please rate the quality, helpfulness, level of details, and relevance of the assistant's answer to the conversation. \\
Use your judgement considering factors such as informativeness, satisfaction, readability, ease of understanding etc. You can also base your judgement on the naturalness, factuality, correctness, usefulness and objectiveness of the answer. \\
The assistant receives an overall score on a scale of 0 to 5, where a higher score indicates better overall performance. Please output "Score: an integer number between 0 and 5". In the subsequent line, please provide a comprehensive explanation of your evaluation, avoiding any potential bias.    \\
        \\ \bottomrule
    \end{tabular}
    \caption{Prompt for GPT-4 evaluation of generate responses.}
    \label{tab:GPT4_eval}
\end{table*}

\section{Mechanical Turk Setup}
\label{sec:app_mturk}

\begin{figure*}[!htb]
  \includegraphics[width=\textwidth]{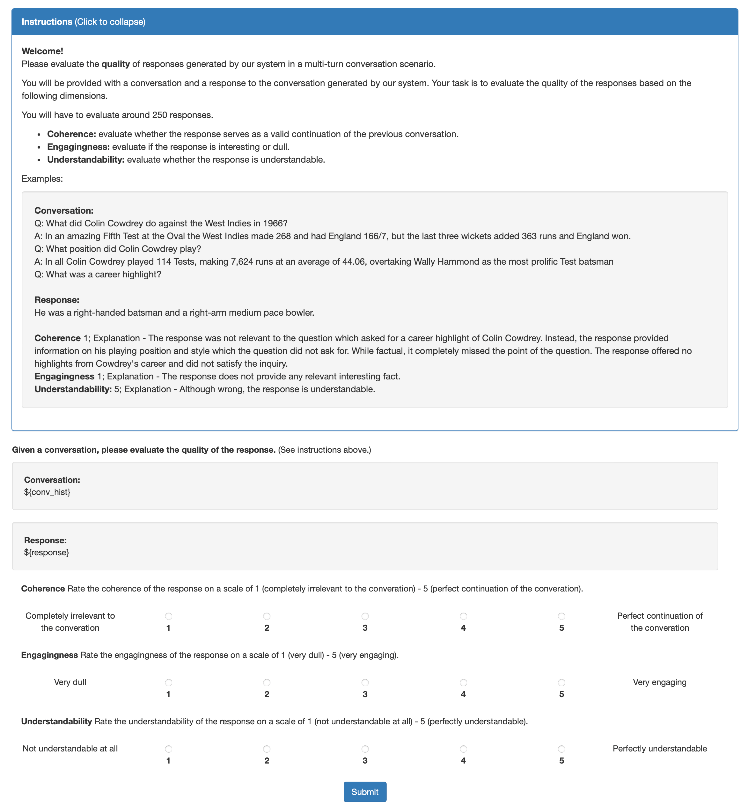}
\caption{Screenshot of the human annotation template for the response quality measurement.}
\label{fig:Mturk}
\end{figure*}

We conduct human evaluations using Amazon Mechanical Turk to determine the quality of the generated answers based on coherence, engagingness and understandability. We give annotators fair compensation. We also use a bonus incentive structure. Every worker who passes the automatic quality checks receives a bonus at the end. In addition, we only consider workers from a country whose main language is English, who has completed 100 or more HITs so far with an acceptance rate of 95\% or higher. Figure~\ref{fig:Mturk} shows the template for evaluating the generated response with respect to the question and conversational history.  In Table~\ref{tab:human_scores_detailed}, we provide the scores of generated response along the dimensions of coherence, engagingness and understandability as rate by the human annotators.

\begin{table*}
    \centering
    \begin{tabular}{ccccc}
    \toprule
        \textbf{Benchmark} & \textbf{Model} & \textbf{Coherence} & \textbf{Engagingness} & \textbf{Understandability} \\ \midrule
           \multirow{4}{*}{QReCC}     & \multirow{1}{*}{\texttt{Mistral-7B-Instruct-v0.2}$^-$} &  4.26& \textbf{3.17}& 4.78 \\ 
     & \multirow{1}{*}{\selfrag{}$^\dagger$}  &3.91& 2.21& 4.52 \\ 
&   \multirow{1}{*}{\selfragorg{}$^\dagger$} & 4.26& 2.3& 4.72 \\
       &  \multirow{1}{*}{\selfragmult{}$^\dagger$} 
 &\textbf{ 4.75}& 2.88& \textbf{4.89}  \\\midrule
   \multirow{4}{*}{Ultrachat}     & \multirow{1}{*}{\texttt{Mistral-7B-Instruct-v0.2}$^-$}   & 4.20& 3.20& 4.70 \\ 
     & \multirow{1}{*}{\selfrag{}$^\dagger$}  & 4.03& 2.63& 4.59 \\ 
&   \multirow{1}{*}{\selfragorg{}$^\dagger$} & 4.39& 2.55& 4.72 \\
       &  \multirow{1}{*}{\selfragmult{}$^\dagger$} 
&\textbf{ 4.81}& \textbf{3.20}& \textbf{4.85} \\\midrule
 \multirow{4}{*}{MT-Eval}     & \multirow{1}{*}{\texttt{Mistral-7B-Instruct-v0.2}$^-$}   &  4.13& 3.08& 4.62 \\ 
     & \multirow{1}{*}{\selfrag{}$^\dagger$}  & 4.08& 2.64& 4.58\\ 
&   \multirow{1}{*}{\selfragorg{}$^\dagger$} & 4.38& 2.63& 4.71 \\
       &  \multirow{1}{*}{\selfragmult{}$^\dagger$} 
& \textbf{4.77}& \textbf{3.23}& \textbf{4.81} \\\bottomrule
    \end{tabular}
    \caption{Detailed result of human annotation scores along different dimensions of generated responses by different models.}
    \label{tab:human_scores_detailed}
\end{table*}

\section{Examples of Generated Responses}
\label{sec:examples_resp}
In Table~\ref{tab:improvements}, we provide an example where \selfragmult{} comprehends longer conversational contexts better than its single-turn counterpart.
\begin{table*}
\tiny
    \centering
    \begin{tabular}{p{2cm}p{8cm}p{2cm}p{2cm}}
    \toprule
        \textbf{Type} & \textbf{Conversation History} & \selfragorg{} &   \selfragmult{} \\ \midrule
        \multirow{3}{2cm}{I. Better processing of conversation history}  & Q: Who was the father of programming? A: Ada Lovelace is often credited as the 'father (or rather, Mother) of programming' or the first computer programmer. Ada was an English mathematician and writer who worked with Charles Babbage on the design of his proposed mechanical general-purpose computer, the Analytical Engine. She is known for writing the first algorithm intended to be processed by a machine, specifically the Analytical Engine. Her work on the engine, including her notes on the method of using the machine to calculate Bernoulli numbers, is considered the first published algorithm and has led her to be recognized as the world's first computer programmer. Q: When was that algorithm written? & She was born in 1815. Her father was the poet Lord Byron. She was a gifted mathematician and wrote an algorithm for the first mechanical computer, the Analytical Engine & Ada Lovelace's algorithm was written between 1842 and 1843.\\ 
        & & & \\
        & Q: Where did the i have a dream speech happen? A: I Have a Dream was delivered from the steps of the Lincoln Memorial in Washington, D.C. Q: Who made the i have a dream speech? A: I Have a Dream is a public speech that was delivered by American civil rights activist Martin Luther King Jr. Q: how big was the audience for the i have a dream speech? A: I Have a Dream is a public speech that was delivered to over 250,000 civil rights supporters. Q: Qass the speech important? & Martin Luther King Jr. made the I Have a Dream speech from the steps of the Lincoln Memorial in Washington, D. C & Yes, the I Have a Dream speech was a pivotal moment in American history and is considered to be one of the most important speeches in American history\\ \midrule
       II. Passage provided in question  & Q: \textbf{Here is a piece of text:} ShadowParty – a supergroup consisting of Devo, New Order and Verve members – have announced their first ever gig. The band is made up of Josh Hager, currently playing guitar and keyboards in Devo, New Order bassist Tom Chapman and guitarist Phil Cunningham and Devo drummer Jeff Friedl. Speaking of the band’s origins, Chapman says it was due to him feeling a bit “restless” after moving to Boston due to family commitments. \textit{\textbf{The four piece have recorded an album, with guests including Denise Johnson (Primal Scream) and The Verve’s guitarist Nick McCabe.}} It’s said to have \textbf{\textit{a synth-pop vibe, with strings and guitars}}. They’ll be making their live debut on May 18th at 1930s cinema, The Plaza, as part of The Charlatans North By Northwich. The gig is free and the show will start around 8pm. You can see the poster below. Last year, Devo were given their own set of emojis. Devo formed in 1973 and released debut album ‘Q: Are We Not Men? A: We Are Devo!’ in 1978. Their last full-length LP ‘Something for Everybody’ was released in 2010. Many of the new emojis reference memorable moments from videos for Devo hits like ‘Whip It’ and ‘(I Can’t Get No) Satisfaction’ as well as other imagery unique to the band.  Based on the text material above, generate the response to the following quesion or instruction: \textit{Can you describe the musical style of ShadowParty and their upcoming debut album?}  & I'm not sure. &  ShadowParty's upcoming debut album is said to have \textbf{\textit{a synth-pop vibe with strings and guitars}}. The band has \textit{\textbf{recorded an album with guests including Denise Johnson (Primal Scream) and Nick McCabe (The Verve)}}. \\  
       & & & \\
       & Q: MILAN (Reuters) - Italian luxury yacht builder Ferretti is preparing for a return to the Milan stock exchange it abandoned more than 15 years ago, a source close to the matter said on Thursday. The group was delisted in 2003 and acquired by private equity firms before being bought by Chinese industrial conglomerate Weichai Group in 2012. \textit{Under the Chinese ownership, the maker of Riva and Pershing yachts has paid off its debt and returned to profit in 2016.} \textit{According to the most recent results available on the company’s website, Ferretti said it had a ``production value'' of 623 million euros (\$709 million) in 2017. The website said 2017 core profit was up 11 percent at 59 million euros, with net profit jumping 71 percent to 24 million euros.} Owner Weichai Group believes that a new listing would improve the company’s reputation and give it the visibility a luxury group deserves, the source said, adding that timing has yet to be decided and will depend on market conditions. The source said the process is at an early stage, with the company having yet to appoint a bank as adviser and no detail available on how much of the company might be floated or at what price. A Ferretti Yachts representative was not immediately available for comment. The group considered a return to the stock market in 2008 but halted the process as the global financial crisis unfolded. The news of Ferretti potentially planning a market comeback were first reported by Italian daily MF. \textbf{Based on the passage above answer:} Has Ferretti made any previous attempts to return to the stock market, and what factors led to the company halting those attempts? A: Yes, Ferretti considered a return to the stock market in 2008, but halted the process as the global financial crisis unfolded. Q: Can you provide more information on Ferretti's financial performance in 2016 and 2017, and how it has improved under Chinese ownership? & Ferretti is preparing for a return to the Milan share market it abandoned more than 15 years ago, a source close to the matter said on Thursday. The group was delisted in 2003 and acquired by private equity firms before being bought by Chinese industrial conglomerate Weichai Group in 2012.   \textit{Under the Chinese ownership, the maker of Riva and Pershing yachts has paid off its debt and returned to profit in 2016} & Absolutely. \textit{According to the most recent financial reports available on Ferretti's website, the company had a `production value' of 623 million euros (\$709 million) in 2017. Core profit for that year was up 11 percent at 59 million euros, with net profit jumping 71 percent to 24 million euros. The company has improved under Chinese ownership, having paid off its debt.} \\\bottomrule
    \end{tabular}
    \caption{Examples of \selfragmult{} improvements in conversational QA over \selfragorg{}.}
    \label{tab:improvements}
\end{table*}

    \end{document}